\documentclass{article}

\usepackage{arxiv}

\usepackage{subcaption}
\usepackage{soul}

\usepackage[ruled,vlined,linesnumbered]{algorithm2e}

\usepackage{tcolorbox} 

\usepackage{booktabs} 
\usepackage{multirow} 
\usepackage{makecell} 
\usepackage{rotating} 
\usepackage{natbib}
\usepackage[hidelinks]{hyperref}
\usepackage{nameref}
\usepackage{amsmath} 
\usepackage{hhline}
\usepackage[nameinlink]{cleveref}

\AddToHook{cmd/appendix/before}{%
    \crefalias{section}{appendix}%
    \crefalias{subsection}{appendix}%
    \crefalias{subsubsection}{appendix}
}

\begin{document}

\title{Leveraging LLMs for Turkish Skill Extraction}

\author{\textbf{Ezgi Arslan İltüzer}~$^{*1}$ \vspace{1em} \textbf{Özgür Anıl Özlü} ~$^{1,2}$  \vspace{1em} \textbf{Vahid Farajijobehdar} ~$^{1}$ \vspace{1em} \textbf{Gülşen Eryiğit} ~$^{1,2}$  \\
\normalfont{\small $^{1}$ Kariyer.net, R\&D Center} \\
\normalfont{\small $^{2}$ Istanbul Technical University	Artificial Intelligence and Data Engineering Department}
}

\maketitle

\begin{abstract}
Skill extraction is a critical component of modern recruitment systems, enabling efficient job matching, personalized recommendations, and labor market analysis. 
Despite Türkiye's significant role in the global workforce, Turkish, a morphologically complex language, lacks both a skill taxonomy and a dedicated skill extraction dataset, resulting in underexplored research in skill extraction for Turkish. This article seeks the answers to three research questions: 1) How can skill extraction be effectively performed for this language, in light of its low resource nature? 2)~What is the most promising model? 3) What is the impact of different Large Language Models (LLMs) and prompting strategies on skill extraction (i.e., dynamic vs. static few-shot samples, varying context information, and encouraging causal reasoning)? The article introduces the first Turkish skill extraction dataset and performance evaluations of automated skill extraction using LLMs. The manually annotated dataset contains 4,819 labeled skill spans from 327 job postings across different occupation areas.

The use of LLM outperforms supervised sequence labeling when used in an end-to-end pipeline, aligning extracted spans with standardized skills in the ESCO taxonomy more effectively. The best-performing configuration, utilizing Claude Sonnet 3.7 with dynamic few-shot prompting for skill identification, embedding-based retrieval, and LLM-based reranking for skill linking, achieves an end-to-end performance of 0.56, positioning Turkish alongside similar studies in other languages, which are few in the literature. 
Our findings suggest that LLMs can improve skill extraction performance in low-resource settings, and we hope that our work will accelerate similar research on skill extraction for underrepresented languages.

\end{abstract}


\section{Introduction}
\label{sec:introduction}
Today, recruitment platforms play a vital role in job markets and contain a growing amount of unstructured textual data. 
Skill extraction is a crucial element in the recruitment sector, enabling efficient matching of candidates to job opportunities, streamlining the screening process, providing personalized recommendations, and offering valuable insights into market trends and skill gaps. Practitioners have used the metaphor ``skills as the currency of the job market'' \citep{Skill-currency-OECD}, showing that skills are one of the main cornerstones of human resource (HR) technology, and drive economic and social growth.

\cite{lightcast-skill} defines skills as an individual's ``abilities, knowledge, and expertise'' needed to perform ``specific tasks or activities in an effective way'', dividing them into ``specialized and common'' types.The concept of skill in the domain of human resources is extensive \citep{beaumont1993} and has broad conceptual frameworks \citep{armstrong2023}. Recently, skill extraction has garnered more attention, particularly with the emergence of large language models and the spreading of generative artificial intelligence (AI) in HR \citep{llm-rethinking}.  However, these studies are still in their early stages and require further research. For example  \citet{llm-rethinking} explores the use of dynamic samples in LLM few-shot prompting for skill identification but does not compare it versus using static samples,  as we do in our study.
 Integrating causal reasoning capabilities into LLMs \citep{kiciman2024causal} has emerged as a promising research direction, with significant implications for natural language processing (NLP), however, their applications to real-world scenarios such as skill extraction are under-explored.
 
\begin{figure}[t]
    \centering
    \includegraphics[width=14cm]{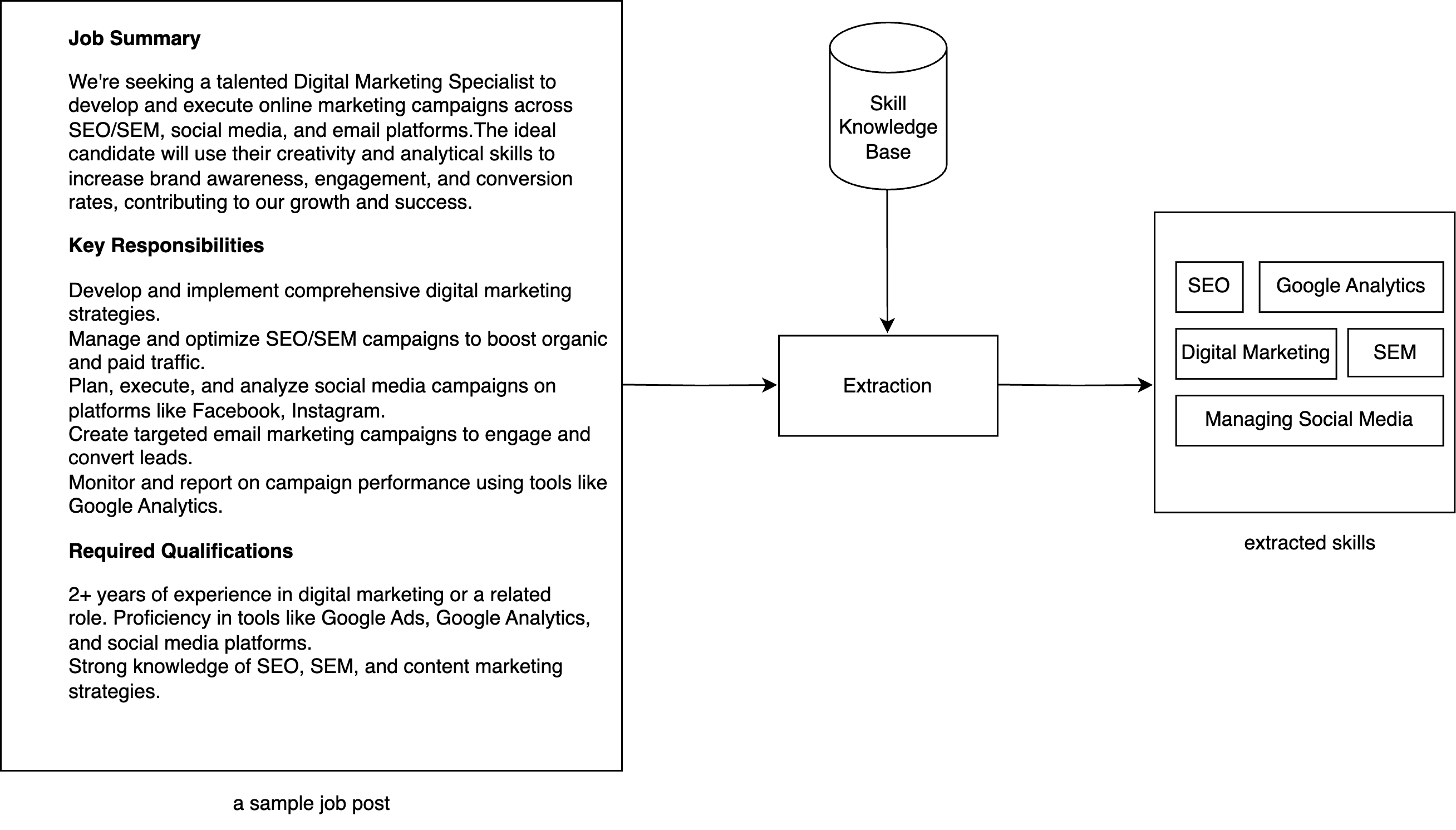}
    \caption{The overview of the skill extraction process. Job postings are treated as unstructured text containing potential skill mentions. The process begins by identifying these potential skills within the text through a skill identification step. Each identified skill candidate is then matched against a predefined skill base to find its equivalent standardized entry. When a successful match is made, the skill mention is linked to the corresponding entry in the skill base, thereby completing the extraction pipeline. This process enables the conversion of raw job posting text into structured skill data suitable for downstream applications.}
    \label{fig:skill-extraction}
\end{figure}

In recruitment, skills are found in various sources such as candidate profiles (CVs and cover letters), job posts (JPs), company career paths, or social media sourcing. Extracting and matching skills from these sources is called ``Skill Extraction'' \citep{senger-etal-2024-deeplearning}, illustrated in ~\Cref{fig:skill-extraction}. Researchers approach skill extraction tasks with diverse algorithms and/or pipelines \citep{senger-etal-2024-deeplearning,nnose,zhang-etal-2024-entitylink} and use different terms such as skill detection \citep{skillgpt}, skill identification \citep{skillgpt}, skill standardization \citep{senger-etal-2024-deeplearning}, skill linking \citep{entitylinking}, skill classification and direct skill classification \citep{senger-etal-2024-deeplearning} for similar or different subtasks.

Based on the terminologies used in the recent studies of \citet{senger-etal-2024-deeplearning} and \citet{entitylinking},  we define the two main stages of our skill extraction pipeline as skill identification and skill linking in our work. Skill identification, which is very similar to some well-known NLP tasks such as named entity recognition and mention detection, extracts textual spans from HR sources potentially referring to some predefined skills without directly linking them to these predefined skills, which is later resolved in the skill linking stage. 
Predefined skills are cataloged and linked in repositories such as ESCO~\citep{esco}, O-NET~\citep{onet2010}, 
\citet{lightcast-skill}, \cite{australia-2023}.
These repositories are referred to as skill knowledge bases (or skill bases for short). Interchangeably, they are also named skill taxonomies or skill ontologies. These resources offer varying levels of structured representation, such as relationships between skills and their definitions, which enhances understanding and analysis.

Skill datasets are pivotal resources for training and evaluating automated skill extraction systems. While there is no strict definition of a skill dataset, such datasets generally consist of labeled textual samples (e.g., JPs, resumes) that illustrate skill mentions and references across different contexts.
References in these datasets link to skill items that are available or previously cataloged in repositories, along with associated descriptions, categories, and more. Unfortunately,  skill datasets are only available for a limited number of languages (i.e., English,  German, Danish, Spanish, and Russian), while others remain resource-poor in this domain.  
The development of skill extraction datasets for different languages is crucial for ensuring AI systems are inclusive, accurate, and capable of supporting global workforce needs across diverse cultural and linguistic contexts.
 
 Turkish, being the focus language of this article, is one of the languages that currently lacks a skill extraction dataset. This gap is particularly significant given that Türkiye's workforce, with its strategic location, growing digital sector, skilled labor, and alignment with international labor standards, contributes significantly to the economic dynamics of both Europe and Asia. It supports global trade, industry, and technological advancement, highlighting the critical need for job market analysis and skill extraction to enhance workforce development and economic growth. This article seeks the answers to the following main research questions:
 \begin{itemize}
     \item How can skill extraction be effectively performed for the Turkish language, in light of its low resource nature?
     \item{What is the most promising model for skill extraction in the Turkish language, based on the current state of the literature, and how can it be applied to achieve optimal performance?
     \item{What is the impact of different LLM prompting strategies on skill extraction (i.e., dynamic vs. static few-shot samples, varying context
information, and encouraging causal reasoning)}?}
 \end{itemize}
 
 As Turkish is a very rich and agglutinative language in terms of morphology, skill mentions in textual context appear in very different surface forms, increasing the need for custom datasets to study the additional language-specific challenges. 
 To address this gap, this research makes a two-fold contribution: 
1) it introduces the first Turkish skill extraction dataset, and 2) it presents the first automated skill extraction performance evaluations using this dataset. We use LLMs for both skill identification and linking stages. In the skill identification stage, we experiment with BERT-based models and decoder-only LLMs (namely BERTTürk, EuRoBERT, GPT, Claude, Gemma, Qwen, Gemini) for sequence labeling, and in the skill linking stage, we experimented with GPT, Claude, Gemma, Qwen, Gemini,  as a reranker on the top 10 retrieved skills via either fuzzy matching or embedding similarity. Our system provides the first performance results on Turkish skill extraction, achieving an F1-score of 0.64 in the skill identification stage and an average end-to-end skill extraction score of 0.56. While there is still room for improvement, this study positions Turkish skill extraction on par with comparable studies in other languages. Additionally, a separate evaluation of skill reranking was performed to compare GPT-4o (which had yielded the best results among the LLMs for skill reranking stage reranking with and without causal reasoning. The results showed that incorporating causal reasoning improved reranking performance, leading to a 2 percentage points (pp) increase.

The remaining of the article is structured as follows; \Cref{sec:relstu} provides the related work, \Cref{section:dataset} introduces the newly developed Turkish skill extraction dataset, \Cref{sec:method} presents the methodology adopted in this research, covering the process of Skill Identification (\Cref{sec:skillidentification}), and then focusing on Skill Linking (\Cref{sec:skilllinking}), which consists of three sub-components: Multi-Skill Parsing, ESCO Skill Retrieval, and LLMs Reranking of Retrieved Skills. \Cref{sec:experimentalsetup}, \Cref{sec:eval}, and \Cref{sec:results} provide the experimental setup, results, and discussions, respectively. \Cref{sec:error-analysis} provides a detailed error analysis of both skill identification and linking stages. \Cref{sec:conclusion} and \Cref{sec:limitations-future-work} offer the conclusion and discuss limitations and directions for future work. Finally, \nameref{sec:practical-implications} outlines the practical implications of our findings.
 
\section{Related Studies}
\label{sec:relstu}
The most common skill base used for annotating and linking skill data is the European Skills, Competences, Qualifications, and Occupations (ESCO)  multilingual classification \citep{esco}. ESCO is one of the leading and most comprehensive standard skill taxonomies in Europe. It provides descriptions for 3,039 occupations and 13,895 skills associated with these occupations, translated into 28 languages, including all official EU languages as well as Icelandic, Norwegian, Ukrainian, and Arabic (up to 2024, V 1.2) \cite{esco-data}.  

\Cref{tab:datasets} provides a list of notable skill extraction datasets, stating the language in focus, the annotation level, and the number of annotated skill mention spans. We categorize the annotation level of the datasets under two types: Skill Mention (SM) involves the annotation of textual mention spans in documents such as JPs and resumes, whereas Skill Reference (SR) involves the process of associating skill mentions with related unique skill identifiers in skill bases. As may be seen from the table, English datasets are more common than those in other languages like Danish, French, or German. \citet{sayfullina} and \citet{green-etal-2022-development} use the same data source, derived from the publicly accessible Kaggle platform, including job postings from various sectors within the UK job market. \citet{sayfullina} provided a list of skill spans, which were extracted using an exact match from a list of soft skills along with a context window and then annotated using crowd-sourcing to determine whether a span represents a soft skill. Apart from \citet{sayfullina}, FIJO \citep{fijo} contains French JPs annotated for soft skills by domain experts, with entities such as Thoughts, Results, Relational, and Personal. These datasets have not been linked to skill references. \citet{green-etal-2022-development} annotates Skills, Qualifications, Experiences, Occupations, and Domains, without relating to skill references.  SKILLSPAN \citep{skillspan} consists of skill spans and knowledge spans from multiple datasets, annotated through an iterative process involving domain experts following the guidelines. 

\begin{table}[!htb]
\centering
\caption{Overview of widely-used skill extraction datasets. Adapted from \cite{senger-etal-2024-deeplearning}. 
SM: Skill Mention, skill spans are identified and annotated in the text.
SR: Skill Reference, Skill spans are additionally linked to entries in a predefined skill taxonomy or knowledge base.}
\label{tab:datasets}
\begin{tabular}{llll} \hline 

 \textbf{Name} & \textbf{Language}& \textbf{Annotation}& \textbf{Size}       \\ \hline  
 SAYFULLINA \citep{sayfullina}& English& SM & 7411 spans\\ 
 GREEN \citep{green-etal-2022-development}& English&SM & 10,606 spans\\ 
 SKILLSPAN \citep{skillspan}& English&SM &9,633 spans\\ 
 DECORTE \citep{designnegative-decorte}& English&SM - SR  &1,618 spans\\  
 FIJO \citep{fijo}& French&SM - SR&932 spans\\
 KOMPETENCER \citep{zhang-etal-2022-kompetencer}& Danish& SM - SR&920 spans\\
 GNEHM-ICT \citep{gnehm2022} & Swiss-German&SM - SR&10,995 spans\\
 KARIYER (this article)& Turkish& SM - SR&4819 spans\\ \hline
\end{tabular}
\end{table}

Several datasets, such as DECORTE and KOMPETENCER have incorporated ESCO labels into existing datasets through manual matching or automated methods. DECORTE, a variant of SKILLSPAN, manually aligned skills with ESCO to create a gold standard. 
\citet{zhang-etal-2022-kompetencer} annotated skill mention spans using the same guidelines as SKILLSPAN and used annotation labels Experience and Skill within the  Danish dataset KOMPETENCER. ESCO labels were incorporated by querying the ESCO API and calculating the Levenshtein distance between skill mention spans. \citet{gnehm2022} annotated the skills of job posts and used the Swiss Job Market Monitor (SJMM) dataset together with ESCO. 

There are multiple approaches and methods in skill extraction studies ranging from span labeling, binary classification, skill extraction with skill reference like coarse or fine-grained labels  \citep{senger-etal-2024-deeplearning}, and topic modeling \citep{ipmref}.  \cite*{designnegative-decorte} approaches skill extraction as a sentence-level task using different negative sampling strategies. Building on this line of research, \cite{Decorte_2025} introduced ConTeXT-match, a bi-encoder model that integrates contrastive learning with token-level attention for extreme multi-label skill extraction. The model achieves state-of-the-art performance on the Skill-XL benchmark of U.S. job advertisements annotated with ESCO skills, while remaining computationally efficient. It is trained on a LLM-augmented synthetic dataset of automatically generated job-ad sentences proposed in their earlier work on extreme multi-label skill extraction using large language models \citep{decorte2023extreme}, allowing large-scale supervision without manual span-level annotation. Although ESCO is multilingual, both ConTeXT-match and Skill-XL are restricted to English data, and the authors explicitly highlight extending such methods to non-English languages as an important future direction. Our work directly contributes to this line of research by applying skill extraction methods to Turkish, a morphologically rich language not previously addressed in this context. \cite{sayfullina} employed neural network-based approaches, including CNN, LSTM, and Hierarchical Attention Models. \cite{green-etal-2022-development} utilized a CRF-based Entity Recognition model. \citet{skillspan} used SpanBERT for skill identification. \cite{zhang-etal-2022-kompetencer} used sequence-tagging of skill labels with RoBERTa and JobBERT. \cite{gnehm2022} focused on iterative training and annotation using jobBERT-de, tailored for a German dataset. \cite{llmranking} presents a zero-shot system leveraging LLMs for skills extraction from job posts, by both generating synthetic training data to train a supervised skill identification classifier and then reranking the top-n matching results from ESCO using GPT-4. \cite{nnose} have introduced the Nearest Neighbor Occupational Skill Extraction (NNOSE) method, incorporating entity linking into skill extraction. For further references, one may refer to \cite{senger-etal-2024-deeplearning} which provides a detailed survey of skill extraction studies using deep learning methods. As in other fields of text processing, the use of LLMs has also attracted increased interest in the domain of skill extraction. SkillGPT \citep{skillgpt}  is an API service designed for skill extraction that uses LLama for skill identification and then outputs the top-k results from  ESCO  with vector similarity search through an embedding database. \cite{llm-rethinking} shows that  LLMs achieve limited performance for skill identification relative to supervised methods.

Recently, \citet{2024-jobskape} explored the use of synthetic datasets for skill extraction using the ESCO skill base. Their approach employs multi-label classification and in-context learning LLMs, including few-shot prompting techniques.  \citet{matkin2024comparative} conducted a comparative analysis of encoder-based Named Entity Recognition (NER) models and LLMs for skill extraction from Russian job advertisements. Using a labeled corpus of over 5,000 annotated postings, they benchmarked models such as GPT-4o, LLAMA 3, and YandexGPT against encoder-based models like DeepPavlov’s RuBERT and XLM-RoBERTa. Extending the multilingual dimension, \citet{kavas-etal-2025-multilingual} proposed a knowledge graph-based framework to match job advertisements with candidate experiences through skill extraction. Their system integrates: (1) entity linking adapted to the ESCO skill base, (2) extreme multi-label classification using the DSPy model with Chain-of-Thought reasoning, and (3) few-shot in-context learning with LLMs. Extracted skills are normalized with ESCO labels and embedded into a multilingual knowledge graph that captures both hierarchical and cross-linguistic relationships within the skill base. Focusing on soft skills, \citet{gavrilescu2025} introduced a neural classification pipeline grounded in the ESCO skill base. Their method combines TF-IDF-based feature extraction, skill-specific binary classifiers, and keyword relevance ranking using neural weights and permutation importance values. Last but not least, \citet{careerbert2025} presented CareerBERT, a domain-adapted Sentence-BERT model designed to semantically match resumes to job profiles from the ESCO skill base. While not explicitly performing skill extraction, CareerBERT enables implicit skill identification by learning semantic relationships between resumes and job categories defined in the ESCO skill base. Similarly, \citet{2023llm4jobs} introduce LLM4Jobs, a framework for unsupervised occupation extraction and standardization from job postings using LLMs and the ESCO; although it focuses on occupations, the method implicitly supports skill-based reasoning by leveraging ESCO’s integrated skill-occupation structure.
As observed in recent literature, the potential of LLMs to trigger cross-lingual transfer \citep{ranaldi-etal-2024-empowering} makes the use of multilingual LLMs valuable for many tasks. 
In light of recent developments, our study experiments with multilingual models, including both BERT-based models and decoder-only LLMs by applying various prompting strategies for skill identification and subsequently reranking the top-k results retrieved from the ESCO skill base.

Regarding studies for Turkish skill datasets,  \citet{VQA-2019} has conducted projects between 2015  and 2019 (under the Türkiye Qualifications Framework (TQF)) to be compatible with the European Qualifications Framework. The developed framework represents the national qualifications that outline all the qualification principles acquired through vocational, general, and academic education or training programs. While it brings value to education or the vocational system including primary, secondary, or higher education and learning pathways, it does not reflect the required skills for jobs or occupations in the domain of the labor market. This project focused merely on qualifications, which is only one of the four pillars (i.e., skills, competencies, qualifications, and occupations) under ESCO. The size of the initial project included 1362 qualifications in the database, incorporating 594 qualifications under the responsibility of the Ministry of National Education. Although there have been attempts to localize ESCO qualifications for the Turkish language, the ESCO skill classification system is unfortunately not yet available for Turkish. Although the full development of this system is far beyond the scope of this article, our aim is to address the gap (at least for the purposes of this study) by automatically translating the ESCO skill classifications into Turkish and manually reviewing the subset of skills referenced from the created skill dataset. The only available study we could find related to Turkish skill extraction is a master’s thesis from \cite{aslim2023}, which aims to reveal skill mismatches in a field of specialization and quantitatively show how well-selected graduate programs meet the requirements of these occupations.  The study is not at a large scale and shows what skills are required for some of the occupations that are most relevant in the chosen field of specialization.

\section{Dataset}
\label{section:dataset}

This section introduces our dataset under two subsections; \Cref{sec:skillbase} provides preparation steps of our skill base relying on ESCO, and \Cref{section:SkillExtractionDataset} provides the annotation stages of our job posting dataset. The dataset will be available upon acceptance.

\subsection{\textit{Turkish Skill Base}}
\label{sec:skillbase}

Linking skill identification outputs to a skill base is essential for establishing a standardized framework for skill representation. This standardization not only ensures consistency across various applications but also enhances the interpretability of content in Turkish, a highly agglutinative language, by eliminating complexities introduced by suffixes. Such linkage facilitates a more accurate understanding of the textual content while creating meaningful and cohesive connections with other studies. Furthermore, it enables us to position our work within the broader literature, allowing for comparative analysis and contributing to the global discourse on skill extraction.

Most existing skill bases are primarily in English or other languages and there is a significant lack of Turkish resources. In order to link our in-text skill mention spans, firstly we automatically translated ESCO version 1.2.0  using an LLM  ( i.e., ChatGPT). The original dataset contains 13,895 skills in English. After the automatic translation, the translations that are only related to our in-text skill mention spans underwent a manual review and update to ensure accuracy. Although a complete and accurate translation of the entire ESCO database into Turkish is beyond the scope of this study, this translated Turkish skill base enables the comparison of skill linking results with those reported in previous studies.

\subsection{\textit{Turkish Skill Extraction Dataset}}
\label{section:SkillExtractionDataset}

\begin{table}[htp]
\centering
\caption{The area distribution of job postings in the KARIYER dataset.}
\label{fig:areadistribution}
\begin{tabular}{l|ccc}
Area & Number of job postings & Total \% & Number of skill spans \\ \hline
Sales - Marketing & 49 & 14.54 & 676 \\ 
Finance & 37 & 10.98 & 625 \\ 
Technical Services & 24 & 7.12 & 263 \\ 
Service & 17 & 5.04 & 116 \\ 
Information Technology & 14 & 4.15 & 372 \\ 
Quality & 13 & 3.86 & 278 \\ 
Machinery & 13 & 3.86 & 128 \\ 
Administrative Affairs & 13 & 3.86 & 264 \\ 
Human Resources & 13 & 3.86 & 228 \\ 
Production & 12 & 3.56 & 180 \\ 
Retail & 11 & 3.26 & 97 \\ 
Construction & 10 & 2.97 & 179 \\ 
Electrical / Electronics & 10 & 2.97 & 112 \\ 
Project / Business / Product Management & 9 & 2.67 & 197 \\ 
Transportation / Logistics & 8 & 2.37 & 78 \\ 
Warehousing / Distribution & 8 & 2.37 & 74 \\ 
Customer Services & 7 & 2.08 & 87 \\ 
Trade & 7 & 2.08 & 104 \\ 
Medicine / Health & 6 & 1.78 & 64 \\ 
Education & 6 & 1.78 & 51 \\ 
Strategy and Reporting / Consultancy & 5 & 1.48 & 75 \\ 
Operations & 4 & 1.19 & 39 \\ 
Environment / Occupational Safety and Health & 4 & 1.19 & 67 \\ 
Office Support & 3 & 0.89 & 31 \\ 
R\&D (Research and Development) & 3 & 0.89 & 62 \\ 
Tourism / Events / Organization & 3 & 0.89 & 23 \\ 
Other & 28 & 8.31 & 353 \\ 
\end{tabular}
\end{table}

In line with \Cref{sec:relstu}, in this section, we will use the terms SM annotation, which involves the annotation of textual mention spans in our JPs, and SR annotation, which involves the process of associating skill mention spans with related unique skill identifiers in ESCO.

For SM annotation, we labeled the skill mention spans  (hereinafter referred to as ``skill spans'') occurring within 327 job posts collected from a popular job portal in Türkiye (i.e., Kariyer.net). The distribution of job areas represented in these posts is shown in \Cref{fig:areadistribution}. The table highlights a pronounced skew toward the Sales \& Marketing and Finance areas. Areas with fewer than $3$ job postings in the dataset, such as ``Art'', ``Law'', and ``Chemistry'', are grouped together under the row labeled ``Other''. This disproportion is deliberately retained to reflect the authentic distribution observed on the online job platform from which the dataset was obtained. Preserving this natural skew allows us to more effectively assess the practical performance of our approach in real-world application scenarios.

Whereas some skill spans appear individually in the JPs, such as ``programming with Python'' or ``speaking English'', there are instances where multiple skills are mentioned using conjunctions, for example, ``having advanced social and technical skills''. In such cases, the combined skill spans that cannot be separated are tagged as ``multi-skill''. This approach enables the identification of complex skill structures within the dataset.

Labeling data is a challenging task that requires careful planning and iteration. To address these challenges, for the SM annotation, we adopted a process inspired by the MAMA (Model-Annotate-Model-Annotate) cycle introduced by \cite{bunt-mama}. The annotation process was carried out in three distinct rounds with a group of three annotators, which consists of two computer scientists and one product manager with a background in marketing took place, all native Turkish speakers.

In the first round, three annotators labeled 10 job postings each based on the initial set of guidelines. After this phase, inconsistencies and edge cases were reviewed collectively, leading to an updated and more robust set of instructions. In the second round, the same annotators labeled a larger subset of 100 job postings using the revised guidelines. This phase allowed for additional identification of ambiguous cases, which were again discussed to further refine the framework. In the third and final round, each annotator reviewed and cross-checked the labels created by the others, ensuring a thorough examination of the dataset.  Majority voting was used to resolve the conflicts.
This collaborative review process helped resolve any remaining discrepancies and ensured that the final dataset adhered to high-quality standards.
As a result of the SM labeling process, the dataset contains 4,819 skill spans, comprising 4,344 single-skill entries and 475 multi-skill entries.

For the Skill Linking phase (i.e., SR annotation), only the test data was further annotated by mapping extracted skills to their corresponding entries in the Turkish ESCO skill base, where available.  At this stage, a different group of annotators consisting of two computer scientists and one product manager with a background in marketing took place. The computer scientists are native Turkish speakers, while the product manager has professional proficiency in Turkish. All annotators possessed advanced English reading skills and had prior experience with text annotation tasks.
The three annotators independently linked each skill span to the most suitable entry in the Turkish ESCO skill base. To ensure contextual accuracy, annotators had access to the full job posting text as well as the descriptions of ESCO skills. Given the extensive size of the skill base (13,895 entries), displaying all possible matches was impractical. Instead, for each extracted skill span, annotators were presented with the 50 most similar entries/references from ESCO, determined using cosine similarity between text embeddings generated by a transformer-based embedding model\footnote{\url{https://huggingface.co/intfloat/multilingual-e5-large}}. Annotators selected the best match from these 50 candidates or left the field empty if no suitable entry was found.

\begin{table}[t]
\centering
\caption{Inter-annotator agreement scores for the Skill Linking phase (SR annotation).}

\label{tab:agreements}
\begin{tabular}{l|lllll}
               & \textbf{Annotator 1} & \textbf{Annotator 2} & \textbf{Annotator 3} & \textbf{All Annotators} & \textbf{Final} \\ \hline
\textbf{Annotator 1}    & 1.00        & 0.70        & 0.58        &                & 0.89  \\
\textbf{Annotator 2}    & 0.70        & 1.00        & 0.62        &                & 0.80  \\
\textbf{Annotator 3}    & 0.58        & 0.62        & 1.00        &                & 0.64  \\
\textbf{All Annotators} &             &             &             & 0.63           &       \\
\textbf{Final}          & 0.89        & 0.80        & 0.64        &                & 1.00   \\ \hline 
\end{tabular}
\end{table}

To assess annotation quality, we measured inter-annotator agreement using Krippendorff’s Alpha with a nominal distance function, where the distance is defined as $0$ if annotators chose the same label and $1$ otherwise. The pairwise agreement scores between each annotator and the final consensus labels were $0.89$, $0.80$, and $0.64$, while the overall inter-annotator agreement was measured as $0.63$. A detailed breakdown is provided in \Cref{tab:agreements}. In the agreement matrix, the cells at the intersections of ``Annotator 1'', ``Annotator 2'', and ``Annotator 3'' indicate pairwise agreement between individual annotators. The cell at the intersection of the ``All Annotators'' row and column reflects the three-way agreement among all annotators. The ``Final'' column and row show the agreement between each annotator’s initial annotations and the final adjudicated labels, while the intersection of the ``Final'' row and column represents the agreement among the consensus annotations themselves, produced through a collaborative resolution process. Overall, $82\%$ of the annotated spans were successfully linked to an ESCO skill during this phase.

The SR annotation process involved two stages: independent annotation, where annotators initially linked skill spans separately, as detailed above, and consensus review, where they collectively reviewed and refined annotations to establish a consensus ESCO label for each span.  Majority voting was used to resolve any remaining conflicts. During the consensus review, only spans without majority agreement (at least two out of three annotators selecting the same entry) were revisited. The final consensus labels were used to evaluate the performance of the skill linking process.

\section{Method}
\label{sec:method}
\begin{figure}[ht]
\centering
\includegraphics[width=0.8\linewidth]{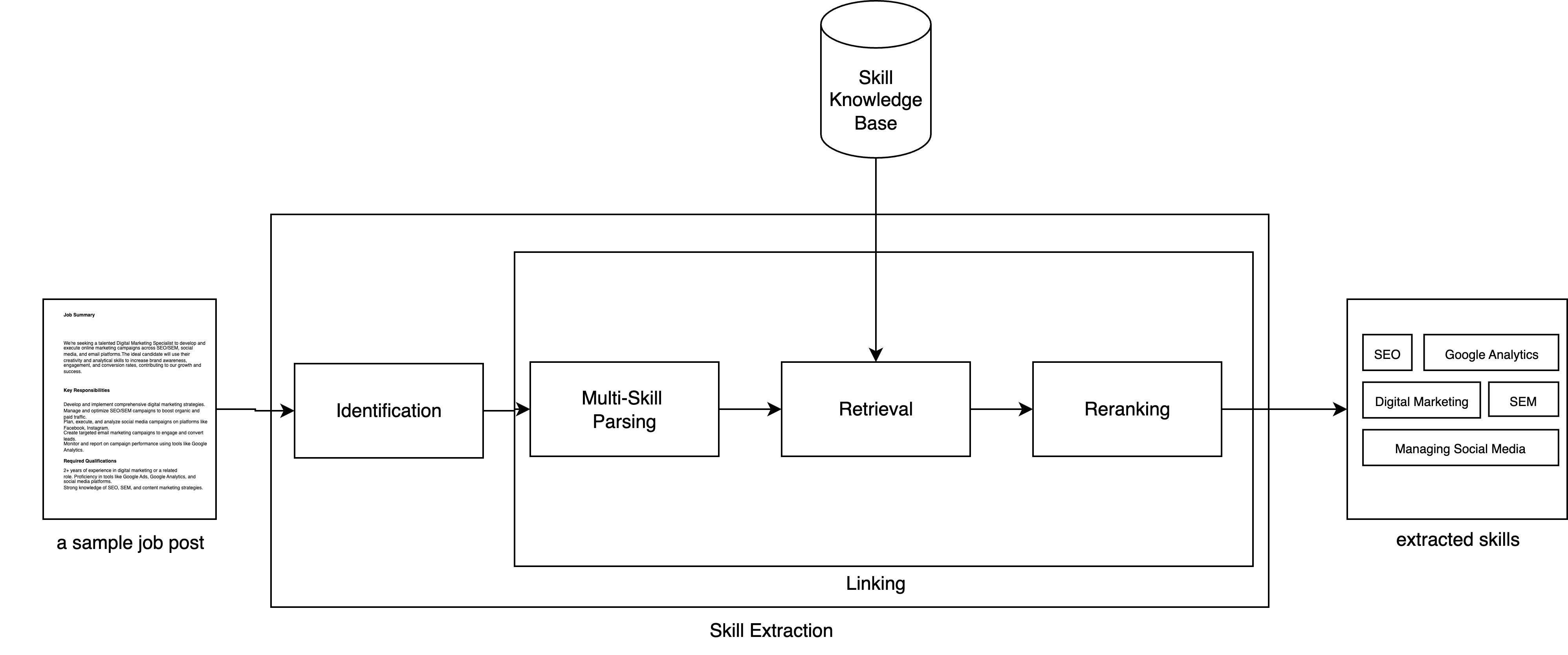}
\caption{High-Level architecture of our Skill Extraction methodology.}
\label{fig:methodology}
\end{figure}

\Cref{fig:methodology} extends the “extraction” box from \Cref{fig:skill-extraction} to provide a high-level view of the overall architecture of our methodology, which follows a multi-stage approach comprising skill identification and skill linking. In addition, \Cref{fig:skill extraction steps} illustrates sample outputs coming from two different retrieval strategies that are going to be introduced below. The remaining of this section is structured as follows: \Cref{sec:skillidentification} provides our skill identification methods, \Cref{sec:skilllinking} provides our skill linking models under 3 subsections which provides our multi-skill parsing strategy (\Cref{section:multiskill}), two different skill retrieval approaches (\Cref{sec:skillretrieval}), and the LLM reranking strategies employed (\Cref{section:llmRerank}).

\subsection{\textit{Skill Identification}}
\label{sec:skillidentification}
This study experiments with two methods for skill identification: 1) Supervised training: fine-tuning an encoder-only language model for sequence labeling of skills and 2) LLM prompting: using a decoder-only language model to directly tag skills in job postings.

Recently, \cite{nguyen-etal-2024-rethinking} experimented with LLM prompting for skill identification. The authors introduced a dynamic few-shot prompting strategy and showed its benefit in comparison to zero-shot prompting. While their work introduces a dynamic strategy for generating similar positive and negative samples, they do not provide a comparison with a static approach where such samples are pre-selected and remain constant throughout the evaluation. To address these gaps, we extended their method by adding standard, statically selected sample instances to our experiments. Instead of relying only on dynamically generated positive and negative samples, we also used a fixed set of pre-selected samples that stayed the same throughout the evaluation. By directly comparing both approaches under the same experimental conditions, we aimed to determine whether the dynamic method offers significant advantages over the static approach in terms of performance and practicality.

Also, LLMs are used to extract skills from job postings. The same job postings from the test set of the dataset outlined in \Cref{section:SkillExtractionDataset} are utilized to ensure a consistent basis for comparison. Each sentence is provided to the LLM with specific instructions to tag skills using $<skill\_start>$ and $skill\_end>$ markers, without altering any other parts of the text (i.e., similar to the NER style prompting in \cite{nguyen-etal-2024-rethinking}); see \Cref{appendix:skill_identification_prompt} for the full prompt.
 
For the static few-shot prompting, an equal number of sentences with and without skills have been selected to use as samples. These sasmples have been fixed for all sentences. For the dynamic few-shot prompting, dynamic prompts have been created for each sentence. As noted by \cite{nguyen-etal-2024-rethinking}, ``to leverage demonstrations that are closely related to each sample, we use a kNN-retrieval approach''. We have generated embeddings for each sentence and retrieved the top k closest sentences in the training set using cosine similarity.

\subsection {\textit{Skill Linking}}
\label{sec:skilllinking}

As explained before, merely identifying skill spans in JPs is insufficient for our objectives. In the skill linking stage, we aim to link the identified spans by our skill identification methods to our ESCO skill base. One should note that the detected spans covering multiple skills should also be handled at this stage. That is why, this section firstly provides a preprocessing stage (\Cref{section:multiskill}) before skill retrieval, and then the two methods employed for skill retrieval: fuzzy matching and embedding similarity (\Cref{section:sentencesim}). Additionally, \Cref{section:llmRerank} provides details about our LLM reranking approaches with and without context.

\subsubsection{Multi-Skill Parsing}
\label{section:multiskill}

During the labeling process, certain data points are tagged as ``multi-skill''. However, these instances are treated as single skills when training the supervised skill identification model. To accurately link the extracted spans to the correct ESCO skills, it is necessary to first detect and split multi-skill spans. 
Given the availability of labeled data, we train a supervised classification model using Support Vector Machines (SVM) \citep{svm} specifically for this task. To represent spans, we use a Bag-of-Words (BoW) approach, where each labeled span is transformed into a vector using count-based encoding. This encoding records the frequency of each word within the span, providing a simple yet effective representation. We use the CountVectorizer implementation from scikit-learn \citep{scikit-learn}, applying a basic whitespace-based tokenization without additional preprocessing steps such as stemming, stopword removal, or subword tokenization. Since the dataset is relatively small, we retain the full vocabulary without dimensionality reduction. This approach ensures interpretability while maintaining the necessary granularity for classification.

\begin{figure}[t]

    \centering
    
    \subfloat[Using Fuzzy Match Based Retrieval]{
      \includegraphics[width=\textwidth]{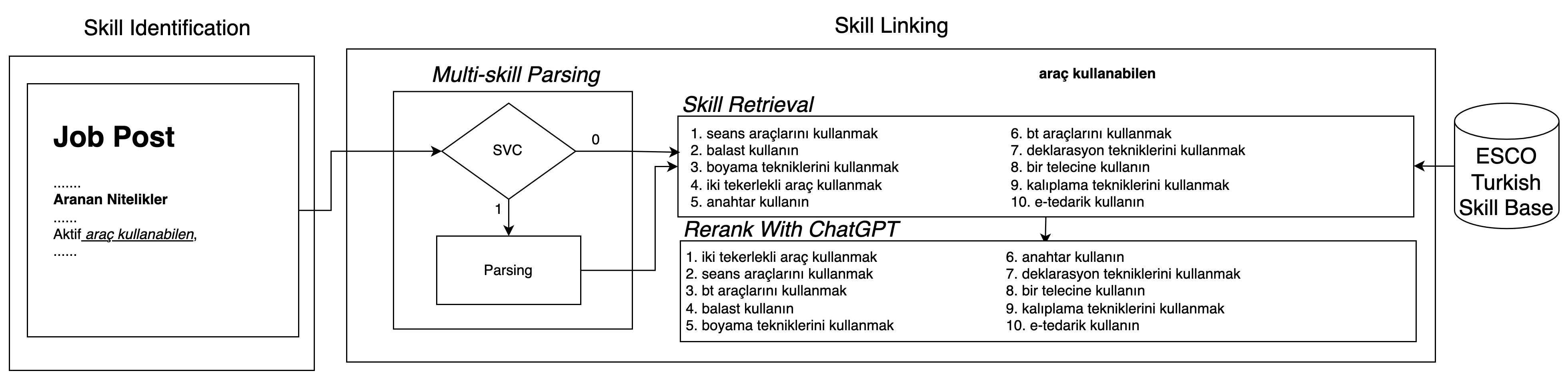}
      \label{fig:skill extraction steps fuzzy}
    }\qquad
    \subfloat[Using Embedding Based Retrieval]{
      \includegraphics[width=\textwidth]{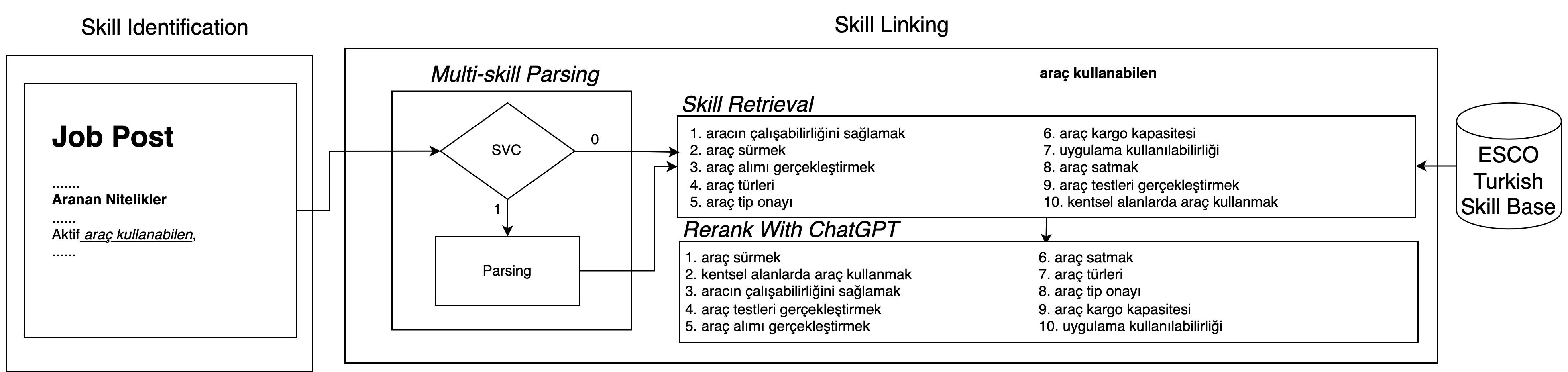}
      \label{fig:skill extraction steps embedding}
    }
    \caption{An instance of original skill span: ``araç kullanabilen''.  (English literal translation: ``able to use vehicle'' meaning ``able to drive''\protect\footnotemark) During the skill identification phase, skill spans in job postings are detected. In the skill linking phase, it is checked whether the skill span is a multiskill, and if so, it is parsed. In the skill retrieval phase, the 10 most similar skills are retrieved from the ESCO skill base, which has been translated into Turkish. These skills are then reranked using an LLM. The main factor influencing variations in similar skills is the difference in skill retrieval methods. It has been observed that the (b) embedding-based method, which can capture semantic similarity, produces more meaningful results than (a) fuzzy match.  }
    \label{fig:skill extraction steps}
\end{figure}

\footnotetext{``able to drive'' may have different meanings in different contexts. For example, when prefixed with a term like ``BT'' (as seen in number 6 of the fuzzy-match retrieval results in Figure 2a), its meaning shifts to ``able to use IT tools''.}

In our workflow, spans produced by the skill identification model are first processed by the classifier to determine whether they contain multiple skills or not. If a span is classified as multi-skill, a rule-based splitting function is applied to generate all possible skill combinations based on predefined delimiters. To handle multi-skill spans, we consider two key scenarios:

    1. If one side of the delimiter contains more words than the other, the remaining words—excluding the one closest to the delimiter—are distributed across both sides of the delimiter. For example, a skill like ``etkili ve anlaşılır iletişim'' (\textit{effective and clear communication}) is identified as a multi-skill and is split into two distinct skills: ``etkili iletişim'' (\textit{effective communication}) and ``anlaşılır iletişim'' (\textit{clear communication}).
    
    2. If both sides of the delimiter have the same number of words, each word farthest from the delimiter is distributed to both the beginning and the end of the span. For example, the multi-skill ``İngilizceyi iyi ve güzel konuşma'' (\textit{English accurately and eloquently speaking}) generates possible outputs such as ``İngilizceyi iyi konuşma'' (\textit{English accurately speaking}), ``İngilizceyi güzel konuşma'' (\textit{English eloquently speaking}), ``İngilizceyi iyi'' (\textit{English accurately}), ``İngilizceyi güzel'' (\textit{English eloquently}), ``iyi konuşma'' (\textit{accurately speaking}) and, ``güzel konuşma'' (\textit{eloquently speaking}). Since the valid outputs will be linked to ESCO skills in the next step, generating all possible combinations and filtering out the nonlinkable ones has been seen to be more effective.

\subsubsection{ESCO Skill Retrieval}
\label{sec:skillretrieval}

The following two approaches have been used to retrieve the most similar ESCO labels to the identified skill spans.

\textbf{Fuzzy Matching}\label{section:fuzzy}: 
Fuzzy matching algorithms are commonly employed to match strings that are similar but not identical. These algorithms work by calculating the similarity between two strings and assigning a score based on their closeness. One common method is the Levenshtein distance, which measures the number of single-character edits required to change one word into another. By applying this algorithm, we were able to generate a list of the most similar ESCO skills for each skill extracted by our skill identification models.

\textbf{Embedding Similarity}\label{section:sentencesim}:
Embedding similarity algorithms are designed to evaluate the similarity between two sentences. These algorithms convert sentences into vector representations and produce a similarity score that reflects how closely the sentences are related in terms of content. By implementing this method, we have generated a list of the most semantically similar ESCO skills corresponding to each span identified by our skill identification models. 

\Cref{fig:skill extraction steps} outlines the steps involved in the skill extraction process, which is divided into two main stages: Skill Identification, and Skill Linking. The process begins with analyzing a job post to identify skill spans, such as keywords or phrases indicating required skills (e.g., ``araç kullanabilen'' (\textit{able to drive}). These skill spans are then passed to the Multi-Skill Parsing module, where the support vector classifier processes the extracted spans. In the Skill Retrieval step, relevant skills are retrieved based on the extracted spans and mapped to candidate ESCO labels. Finally, the retrieved skills are reranked using ChatGPT, to refine the ordering and ensure the relevance of the final skill list. \Cref{fig:skill extraction steps fuzzy} and \Cref{fig:skill extraction steps embedding} provide sample outputs retrieved with both fuzzy match based and embedding based retrieval strategies respectively. The example illustrates the morphological complexity of the Turkish language, where the lemmas of words forming the identified skill span (i.e., ``araç'' (\textit{vehicle}) and ``kullan'' (\textit{use}) ) occur under several different surface forms (\Cref{fig:skill extraction steps fuzzy}) such as ``araçlarını, araç''(\textit{his/her vehicles, vehicle}) and ``kullanabilen, kullanmak, kullanın'' (\textit{(who is) able to drive, driving, (imperative) drive)}. The figure shows that fuzzy match based retrieval brings labels that are lexically similar whereas embedding based retrieval  (\Cref{fig:skill extraction steps embedding}) brings labels that are semantically similar such as ``araç sürmek'' (\textit{literally: vehicle drive, meaning: to drive}). 

\subsubsection{LLM Reranking of Retrieved Skills}
\label{section:llmRerank}

We hypothesize that performing LLM reranking over retrieved results will yield higher performance results compared to selecting only the top-1 match. Initially, fuzzy matching and embedding similarity algorithms are used to identify the 10 closest matches between the skill spans extracted by the skill identification models and the labels in the ESCO database.

To refine the matching of identified skill spans with ESCO labels, we conducted 4 different trials. As the baseline model (W/O Rerank), the most similar match is selected based on skill retrieval methods without any reranking. As the next step, an LLM is used to rerank the top 10 similar items generated by skill retrieval methods, allowing for a more precise selection of the most appropriate skill. Three prompting strategies are experimented with for this reranking stage: first, only the identified skill span and retrieved ESCO labels were provided (w/ RerankKey); second, contextual information was incorporated into the prompt by also including the sentence in which the span appears (w/ RerankContext);  third, in addition to the second approach, the descriptions of the most similar ESCO labels were also included (w/RerankContext-Description). These trials aimed to evaluate the influence of reranking strategies and contextual enrichment on the performance of skill linking. 

Additionally, we explore whether prompting the LLM to justify its ranking decisions can further improve performance. Building on our existing reranking strategies, we introduce two reasoning-based variations. In the first (\Cref{appendix:skill_linking_prompt-reason_rerank}, Reason + Rerank Prompt), the LLM is prompted to compare each retrieved ESCO label with the identified skill span, explicitly explaining the similarities and differences before reranking. In the second (\Cref{appendix:skill_linking_prompt-causal_reason_rerank}, Causal Reason + Rerank Prompt), 
the LLM is prompted to reason why each retrieved ESCO skill is relevant to the posted job posts and how hiring a candidate with that ESCO skill —rather than the identified skill span— would impact the fulfillment of the expected role's requirements from an employer's perspective. By incorporating causal reasoning, we move beyond simple ranking based on textual similarity and encourage the model to engage in counterfactual evaluation. Instead of treating reranking as a static matching problem, this approach prompts the LLM to actively consider the functional consequences of hiring a candidate with each retrieved skill. We hypothesize that this approach will allow the system to better account for real-world applicability, rather than relying solely on linguistic similarity.

\section{Experiments \& Results}

This section is organized as follows: \Cref{sec:experimentalsetup} provides the experimental setup, \Cref{sec:eval} details the evaluation metrics and methodologies applied to assess both individual components and the end-to-end pipeline, and finally, \Cref{sec:results} presents and discusses the results.

\subsection{\textit{Experimental Setup}}
\label{sec:experimentalsetup}

Our Turkish skill extraction dataset (\Cref{section:SkillExtractionDataset}) consists of spans labeled with IOB tags to ensure precise skill detection. The dataset is divided into training, validation, and test subsets, with careful consideration of the major occupation categories \citep{isco} to maintain a balanced representation. Specifically, if a category appears in only one job posting, it is excluded from the test and validation sets to prevent under-representation. For occupation categories with more than six job postings, 20\% of the data is allocated for testing, 20\% for validation, and the remaining 60\% for training. As a result, 18.8\% of the total skill spans were placed in the test set.

For the skill identification step, we fine-tune two transformer-based models on our dataset using the Named Entity Recognition (NER) objective. The first is a Turkish BERT model\footnote{\url{https://huggingface.co/akdeniz27/bert-base-turkish-cased-ner}} available on Hugging Face. The second is EuroBERT-210m \citep{eurobert}, a multilingual model pre-trained on various European languages, including Turkish. Fine-tuning details and hyperparameter settings are provided in \Cref{sec:transformerfinetuning}.

For static and dynamic prompting strategies in skill identification, we use several large language models (LLMs), including OpenAI’s GPT-4o \citep{gpt4o} (i.e., version: gpt-4o-2024-05-13), GPT-4.1\footnote{\url{https://platform.openai.com/docs/models/gpt-4.1}}, Claude 3.7 Sonnet\footnote{\url{https://www.anthropic.com/news/claude-3-7-sonnet}}, Gemini 2.5 Flash\footnote{\url{https://deepmind.google/models/gemini/flash/}}, Gemma 3 \citep{gemma3} (i.e., gemma-3-27b-it), and Qwen3 \citep{qwen3} (i.e., Qwen3-32B). All generations are deterministic to ensure consistent evaluation. Prompting configurations are detailed in \Cref{sec:promptconfig}. For dynamic prompting, a k-nearest neighbors (kNN) model selects in-context examples from the training data. (see \Cref{sec:dynamicprompt}) Model performance is evaluated using CoNLL-F1 and partial MUC F1 scores.

For multi-skill parsing, we employ the Support Vector Classifier (SVC) implementation from scikit-learn \citep{scikit-learn}, using a Radial Basis Function (RBF) kernel.

For skill retrieval, we apply both fuzzy-match and embedding-based similarity methods. The former uses the Rapidfuzz library \citep{rapidfuzz}, while the latter employs a sentence-transformer model with cosine similarity. Detailed configurations of these methods are described in \Cref{sec:appendixskillretrieval}. In certain configurations, we further apply LLM-based reranking using the same set of models.(see \Cref{sec:appendixllmreranking})

\subsection{Evaluation}
\label{sec:eval}
The end-to-end system developed for skill extraction consists of several key components. Among these, the Skill Identification and Skill Retrieval parts play a critical role in the overall pipeline. To gain deeper insights into their impact on the system’s performance, we first evaluate these components in isolation before assessing the full pipeline.

\subsubsection{Skill Identification}
To evaluate the performance of our skill identification models, we use two widely adopted Named Entity Recognition (NER) evaluation metrics: the CoNLL-F1 score \citep{conll} and the SemEval MUC Partial score \citep{partial}. The MUC Partial score is particularly useful in this context, as partial matches of skill spans can still lead to correct ESCO ID links.

For LLM-based prompting models, we convert their outputs into IOB-tagged sequences \citep{Ramshaw1999} based on the original dataset, enabling a direct comparison with our supervised model. However, as noted by \cite{nguyen-etal-2024-rethinking}, LLMs often struggle to preserve exact texts from the input sentence. Instead, they tend to generate the most probable sequence, which can introduce errors such as incorrect punctuation, typos, or altered texts, complicating evaluation. To mitigate these issues, we manually post-processed the outputs to align them more closely with the gold standard dataset.

\subsubsection{\textit{Multi-skill Parsing and Skill Retrieval}}
\label{sec:eval:skillretrieval}
    
Although several studies explore skill extraction (see \Cref{sec:introduction} and \Cref{sec:relstu}), research specifically evaluating the end-to-end performance of both skill identification and skill linking remains less common \citep{designnegative-decorte,decorte2023extreme,llmranking}. These studies often rely on sentence-level evaluation using RP@K and MRR scores. However, since skill linking requires selecting a single correct skill label from a taxonomy, we believe that HitRate@k is a more appropriate metric than R-Precision@K.

To fairly assess our end-to-end system, we evaluate the multi-skill parsing and retrieval stages (i.e., the first and second stages of skill linking) separately from both skill identification and each other. The performance of the multi-skill parsing classifier is measured in isolation using F1 scores. For the retrieval stage, we compute HitRate@k over gold-standard mention spans, considering only those that can be linked to ESCO. HitRate@k quantifies how often the correct ESCO ID is retrieved when using human-labeled skill spans as queries. This evaluation also helps establish an upper bound for the next stage (LLM reranking) by varying $k$, the number of matches passed forward.

\begin{equation}
HitRate@K = \frac{\sum_{i=1}^{N} \mathbf{I}(\text{correct skill in top } k_i)}{N}
\end{equation}

Where $N$ is the total number of skills in the gold test set that are linked to an ESCO label, and
$\mathbf{I}$ is an indicator function, which is $1$ if the correct skill for the i-th skill span appears in the top $k$ predictions, and $0$ otherwise.

\subsubsection{End-to-End Skill Extraction}
\label{sec:end-to-end-eval}

\begin{algorithm}[ht]
\SetAlgoLined
\KwIn{gold: Gold ESCO skills on job posts\\
      \quad \quad \quad predict: Predicted ESCO skills on job posts}
\KwOut{result: Evaluation score}
\BlankLine
\textbf{Initialize:} correct\_matches $\gets 0$ \tcp*[h]{Count of correct matches}\\
\ForEach{job\_ID $j$ in gold}{
    \textbf{Count:} gold\_skills $\gets$ Skills in gold[j] \\
    \textbf{Count:} pred\_skills $\gets$ Skills in predict[j] \\
    \ForEach{skill $s$ in gold\_skills}{
        \If{$s$ is in pred\_skills}{
            correct\_matches $\gets$ correct\_matches $+$ 1
        }
    }
}
result $\gets \frac{correct\_matches}{\text{Number of ESCO skills in gold}}$ \\
\Return result
\caption{End-to-End Skill Extraction Evaluation}
\label{alg:evaluate}
\end{algorithm}

As explained above, our skill extraction framework consists of two components; skill identification and skill linking. In order to measure our end-to-end performance on skill extraction from JPs, we provide an algorithm (i.e, \Cref{alg:evaluate}) that evaluates the performance of skill extraction by comparing the predicted skills against the gold-standard ESCO skills for each job. The process begins by initializing a counter, $correct\_matches$, to track the number of correctly predicted skills. For each job, the algorithm retrieves the skills labeled in the gold-standard dataset $gold\_skills$ and the skills predicted by the model $pred\_skills$ (i.e., positioned at top 1). It then iterates over each skill in the gold-standard skills list and checks if that skill is also present in the predicted skills list. For each match found, the $correct\_matches$ counter is incremented. Finally, the algorithm computes the evaluation score by dividing the total number of correct matches $correct\_matches$ by the total number of skills present in the gold-standard dataset. 

To summarize, this evaluation compares the list of manually annotated and linked skills with those predicted by the end-to-end system for each job advertisement, rather than comparing every individual skill label and corresponding prediction. This approach accounts for differences in predicted spans, which may cause some skills to have no predictions and non-skills to have predictions, and the multiple skill candidates generated through multi-skill parsing.

\subsubsection{Human Evaluation}
\label{sec:humeval}
To assess the real-world quality of the proposed skill extraction system, a structured human evaluation was conducted. The objective was to analyze the validity and relevance of skills extracted automatically from a sample of job advertisements. The evaluation was performed in collaboration with a team of six expert hiring consultants\footnote{They are employed by WeHire \url{https://www.wehire.consulting/}, a recruitment / hiring service / consultancy brand that operates under the  Kariyer.net recruitment platform.}. Their expertise in recruitment and candidate assessment ensured that our human evaluation judgments about skill validity and priority reflected real-world hiring practices.
The evaluation was designed to verify two primary aspects of the system’s output:
1) Skill Validity: Whether items extracted from the job advertisement represent valid, real-world skills.
2) Priority Classification: The importance of a valid skill for the specific role advertised.
An annotation interface was provided in the form of a shared spreadsheet. Experts reviewed the extracted skills for each job advertisement and recorded their assessments, classifying them according to the legend below and adding explanatory notes..

The hiring experts used a color-coded system to classify the priority of each skill extracted from the job advertisement:
\vspace{-0.3em}
\begin{itemize}\setlength\itemsep{-0.3em}
    \item Red: A critical, ``must-have'' skill mentioned in the job advertisement.
\item Blue: An optional or preferred skill that adds value but is not mandatory.
\item Gray: The perception of skill is not clear in Turkish labor market even if listed in ESCO.
\item Yellow: Experts provided written feedback to highlight errors or omissions in system outputs, or opinions regarding the output.
\end{itemize}

The human evaluation of the samples was carried out in two distinct phases across three online sessions, with a total duration of six hours. The qualitative assessment commenced with independent evaluations by subject-matter experts possessing domain-specific knowledge relevant to the analyzed job advertisements (a total of 23, categorized according to this criterion). Subsequently, the individual analyses were consolidated through three consensus panel (i.e., online sessions) discussions to ensure interpretive alignment and reliability.

\subsection{Results \& Discussion}
\label{sec:results}

Our first experiment set provides the performances of our skill identification stage in terms of CoNLL-F1 and MUC Partial scores with different models (i.e., supervised sequence labeling (hereinafter referred to as SSL) via BERT models, and LLMs evaluated under zero-shot and few-shot settings, using both static and dynamic prompting methods) introduced in \Cref{sec:skillidentification}. \Cref{tab:skillidentification} shows that our results align with the findings of \citet{nguyen-etal-2024-rethinking}, who demonstrated that dynamic prompting generally improves LLM performance over zero-shot methods. Extending their analysis, our experiments also include static few-shot prompting and show that dynamic prompting outperforms both zero-shot and static few-shot methods in terms of CoNLL-F1 scores. With the strict span-boundary requirements of this evaluation,  
LLMs evaluated under the zero-shot prompting approach perform poorly, achieving F1 scores between 0.15 and 0.26.

However, the situation changes with our further evaluation using MUC Partial scores. As discussed earlier, since partial matches of skill spans can still lead to correct ESCO ID links, we believe MUC Partial analyses are still valuable for the task in focus.
A deeper analysis (\Cref{tab:skillidentification}, 4$^{th}$ and 8$^{th}$ columns) reveals that 
different LLMs react differently to dynamic prompting in terms of partial matching; while trying to improve their exact match performances, their partial match capabilities generally tends to drop (e.g., Gemma 0.81 vs 0.69 MUC Partial scores for static two-shot vs dynamic two-shot respectively, and GPT-4o 0.76 vs 0.70. These results suggest that dynamic few-shot prompting improves the LLM’s ability to follow explicit task instructions, thereby enhancing exact span identification and increasing CoNLL-F1 scores. However, this improvement does not extend to the model’s ability to recognize skill mentions that are semantically correct but deviate from the annotated spans, as evidenced by largely unchanged or slightly decreased MUC Partial scores. This indicates that while dynamic prompting boosts syntactic alignment with the task, it may not effectively enhance the model’s utilization of domain knowledge for identifying non-exact, contextually valid skills.

Among the LLMs, the highest MUC Partial score (0.81) is achieved by Gemma, while the best CoNLL-F1 score (0.57) is obtained with Claude.\footnote{\Cref{tab:meanskillidentification} in the Appendix Section provides the mean results for these two models together with the standard errors at the end of three runs. The differences between the performances of the closest-performing models are found to be statistically significant based on Almost Stochastic Dominance \citep{almoststochasticorder} with $\epsilon = 0$.}
Nevertheless, our SSL via BERT model remains the top-performing model overall.\footnote{Multilingual EuroBERT also performs poorer when compared to its monolingual counterpart.}.
Notably, the difference in MUC Partial scores between BERT (0.84) and GPT-4o under the zero-shot setting (0.79) is only 5pp, substantially smaller than the nearly 50pp gap observed in CoNLL-F1 scores. These findings suggest that while LLMs still lag behind in strict span-level identification, they are competitive in capturing semantically relevant spans. Importantly, the performance gap between LLMs and the SSL model narrows considerably with the use of prompting strategies, reducing to just 7 percentage points for CoNLL-F1 and 3 percentage points for MUC Partial, for example, Claude Sonnet 3.7 with dynamic ten-shot prompting achieves a CoNLL-F1 score of 0.77, and Gemma 3 with static two-shot prompting reaches a MUC Partial score of 0.81. \Cref{section:appendixArea} provides a more detailed analysis of our two relatively well-sampled areas, sales-marketing and finance, to further discuss the effect of sample size. Although there is room for improvement, these results place Turkish skill identification on par with similar studies conducted in other languages. For instance, \cite{skillspan} achieved a CoNLL-F1 score of 0.59 for English, while \cite{llm-rethinking} reported an average CoNLL-F1 score of 0.64 for NER models on 4 different languages (i.e., English, French, German and
Danish) with a supervised model.

\begin{table}[ht]
\centering
\caption{Skill Identification results.}
\label{tab:skillidentification}
\begin{tabular}{c|lcc||c|lcc}
\multicolumn{1}{c|}{\textbf{LLM}} &
\begin{tabular}[c]{@{}c@{}}\textbf{Prompt}\\\textbf{~Strategy}\end{tabular} &
\begin{tabular}[c]{@{}c@{}}\textbf{CoNLL}\\\textbf{~F1}\end{tabular} & \begin{tabular}[c]{@{}c@{}}\textbf{MUC}\\\textbf{~Partial}\end{tabular} & \multicolumn{1}{c|}{\textbf{LLM}} & \begin{tabular}[c]{@{}c@{}}\textbf{Prompt}\\\textbf{~Strategy}\end{tabular} &
\begin{tabular}[c]{@{}c@{}}\textbf{CoNLL}\\\textbf{~F1}\end{tabular} & \begin{tabular}[c]{@{}c@{}}\textbf{MUC}\\\textbf{~Partial}\end{tabular}\\ \hline
\mbox{\textbf{BERT}} & - & \textbf{0.64} & \textbf{0.84} & 
    \mbox{\textbf{EuRoBERT}}  & -& 0.47 & 0.64 \\ \hline\hline

\multirow{7}[10]{*}{\rotcell{\mbox{\textbf{GPT-4o}}}} 
& Zero-Shot& 0.15 & 0.79 & 
\multirow{7}[10]{*}{\rotcell{\mbox{\textbf{GPT-4.1}}}} 
& Zero-Shot& 0.19 & 0.63 \\
& Static Two-Shot     & 0.11 & 0.76 &
    & Static Two-Shot     & 0.22 & 0.76 \\
& Static Six-Shot     & 0.14 & 0.78 &
    & Static Six-Shot     & 0.20 & 0.75 \\
& Static Ten-Shot     & 0.16 & 0.79 &
    & Static Ten-Shot     & 0.24 & 0.68 \\
& Dynamic Two-Shot & 0.30 & 0.70 &          
    & Dynamic Two-Shot & 0.20 & 0.64 \\
& Dynamic Six-Shot & 0.31 & 0.73 &          
    & Dynamic Six-Shot & 0.19 & 0.64 \\
& Dynamic Ten-Shot & 0.35 & 0.76 &          
    & Dynamic Ten-Shot & 0.20 & 0.64 \\ \hline
 
\multirow{7}[10]{*}{\rotcell{\mbox{\begin{tabular}[c]{@{}c@{}}\textbf{Claude}\\\textbf{~Sonnet 3.7}\end{tabular}}}} 
& Zero-Shot& 0.26	& 0.62
& \multirow{7}[10]{*}{\rotcell{\mbox{\textbf{Gemma 3}}}} 
    & Zero-Shot& 0.19 & 0.44 \\
& Static Two-Shot     & 0.23	& 0.74 &
    & Static Two-Shot     & 0.18	& \textbf{0.81} \\
& Static Six-Shot     & 0.16 &	0.74 &
    & Static Six-Shot     & 0.25 &	0.77 \\
& Static Ten-Shot     & 0.26 &	0.75 &
    & Static Ten-Shot     & 0.32 &	0.72 \\
& Dynamic Two-Shot & 0.48	& 0.73 &   
    & Dynamic Two-Shot & 0.41 & 0.69 \\
& Dynamic Six-Shot & 0.53 & 0.75 &                 
    & Dynamic Six-Shot & 0.52 & 0.73 \\        
& Dynamic Ten-Shot & \textbf{0.57} & 0.77 &        
    & Dynamic Ten-Shot & 0.51 & 0.74 \\ \hline

\multirow{7}[10]{*}{\rotcell{\mbox{\textbf{Qwen3}}}}  
& Zero-Shot& 0.25	& 0.47
    & \multirow{7}[10]{*}{\rotcell{\mbox{\begin{tabular}[c]{@{}c@{}}\textbf{Gemini}\\\textbf{~Flash 2.5}\end{tabular}}}}
    & Zero-Shot& 0.15 & 0.44 \\
& Static Two-Shot     & 0.17	& 0.71 &
    & Static Two-Shot     & 0.21	& 0.63 \\
& Static Six-Shot     & 0.16 &	0.74 &
    & Static Six-Shot     & 0.19 &	0.77 \\
& Static Ten-Shot     & 0.21 &	0.66 &
    & Static Ten-Shot     & 0.24 &	0.71 \\
& Dynamic Two-Shot & 0.44	& 0.66 &   
    & Dynamic Two-Shot & 0.38 & 0.66 \\
& Dynamic Six-Shot & 0.49 & 0.69 &                 
    & Dynamic Six-Shot & 0.44 & 0.73 \\        
& Dynamic Ten-Shot & 0.51 & 0.72 &        
    & Dynamic Ten-Shot & 0.44 & 0.69 \\ \hline
\end{tabular}
\end{table}

\begin{table}[!htb]
\centering
\caption{Skill Retrieval results on gold-standard skill mentions.}
\label{tab:skilllinkinggoldhitrates}
\begin{tabular}{lcccc} \hline 
\textbf{Skill Retrieval}  & \textbf{HitRate@1} & \textbf{HitRate@3} & \textbf{HitRate@5} & \textbf{HitRate@10}\\ \hline
Fuzzy Matching & 0.21 & 0.31 & 0.37 & 0.46\\
Embedding Similarity & 0.52 & 0.74 & 0.79 & \textbf{0.88} \\ \hline
\end{tabular}
\end{table}

As introduced in \Cref{sec:eval:skillretrieval}, our second set of experiments evaluates the performance of the multi-skill parsing and skill retrieval stages in isolation. First, we assess the effectiveness of our multi-skill parsing classifier, an SVM-based model that determines whether a span represents multiple skills (as explained in \Cref{section:multiskill}). The performance of this classifier is measured as 0.89 F1 with a precision of 0.96, and a recall of 0.83.

Next, we evaluate skill retrieval performance, as summarized in \Cref{tab:skilllinkinggoldhitrates}, which reports the hit rates at different top k retrieved results for our two different retrieval methods: fuzzy matching and embedding similarity. \Cref{tab:skilllinkinggoldhitrates} (column HitRate@10) illustrates the potential upper bound of the system performance in the case of perfect reranking. By computing HitRate@k scores over gold-standard skill annotations, we assess how effectively the system retrieves the correct ESCO IDs using human-labeled skill spans as queries. The results indicate that embedding similarity-based retrieval consistently outperforms fuzzy matching across all HitRate@k metrics. This underscores the effectiveness of embedding based retrieval in capturing relevant skills and providing a strong foundation for the reranking stage. The low scores of fuzzy match based retrieval are due to the morphological complexity of the highly agglutinative Turkish language, as depicted in \Cref{fig:skill extraction steps fuzzy}, which increases the character edit distance lengths between different word surface forms. The highest obtained HitRate@10 result is observed as 0.88.

\begin{table}[!htb]
\centering
\caption{Skill Retrieval + Reranking Results on Gold-Standard Skill Mentions \\{\footnotesize 
Skill Retrieval Approach: Embedding Similarity}
}
\label{tab:rerank-methods-gold-llm}
\begin{tabular}{c|lc||c|lc}
\multicolumn{1}{c|}{\textbf{Reranker}} & \textbf{Reranking Prompt Strategy} & \textbf{HitRate@1} & \multicolumn{1}{l|}{\textbf{Reranker}} & \textbf{Reranking Prompt Strategy} & \textbf{HitRate@1} \\ \hline
 & w/o Rerank & 0.52 & 
     & w/o Rerank & 0.52\\ \hline

\multirow{3}[4]{*}{\rotcell{\mbox{\begin{tabular}[c]{@{}c@{}}\textbf{GPT}\\\textbf{-4o}\end{tabular}}}}
& w/ RerankKey & 0.66 &
    \multirow{3}[4]{*}{\rotcell{\mbox{\begin{tabular}[c]{@{}c@{}}\textbf{GPT}\\\textbf{-4.1}\end{tabular}}}} 
    & w/ RerankKey & 0.65\\
& w/ RerankContext & 0.66 & 
    & w/ RerankContext & 0.63\\
& w/ RerankContext-Description     & 0.66 &
    & w/ RerankContext-Description & 0.28 \\ \hline
    
\multirow{3}[7]{*}{\rotcell{\mbox{\begin{tabular}[c]{@{}c@{}}\textbf{Claude}\\\textbf{Sonnet}\\\textbf{3.7}\end{tabular}}}} 
& w/ RerankKey & 0.65 &      \multirow{3}[8]{*}{\rotcell{\mbox{\begin{tabular}[c]{@{}c@{}}\textbf{Gemma}\\\textbf{3}\end{tabular}}}} 
    & w/ RerankKey & 0.63  \\
& w/ RerankContext & 0.66 & 
    & w/ RerankContext & 0.61\\
& w/ RerankContext-Description     & 0.63 &
    & w/ RerankContext-Description & 0.65 \\ \hline

\multirow{3}[5]{*}{\rotcell{\mbox{\begin{tabular}[c]{@{}c@{}}\textbf{Qwen}\\\textbf{3}\end{tabular}}}}  
& w/ RerankKey & 0.64
    & \multirow{3}[8]{*}{\rotcell{\mbox{\begin{tabular}[c]{@{}c@{}}\textbf{Gemini}\\\textbf{Flash}\\\textbf{2.5}\end{tabular}}}}
    & w/ RerankKey & 0.63 \\
& w/ RerankContext & 0.56 & 
    & w/ RerankContext & 0.59\\
& w/ RerankContext-Description & 0.55 &
    & w/ RerankContext-Description & 0.60 \\ \hline
\end{tabular}
\end{table}

\begin{table}[!htb]
\centering
\caption{Impact of Reasoning-Enhanced Prompts \\{\footnotesize Reranker: GPT-4o, Skill Retrieval Approach: Embedding Similarity, Skill Retrieval + Reranking Results on Gold-Standard Skill Mentions}}
\label{tab:rerank-methods}
\begin{tabular}{lcccc} \hline 
\textbf{Reranking Prompt Strategy}  & \textbf{HitRate@1}\\ \hline
w/o Rerank & 0.52 \\
\hline
w/ RerankKey Prompt  & 0.66  \\
w/ RerankContext Prompt & 0.66 \\
w/ RerankContext-Description Prompt & 0.66 \\\hline
w/  Reason + RerankKey   Prompt & 0.65 \\
w/  Reason + RerankContext Prompt & 0.65 \\
w/  Reason + RerankContext-Description  & 0.65 \\ \hline
w/ Causal Reason + RerankKey   Prompt & 0.68 \\
w/ Causal Reason + RerankContext Prompt & 0.68 \\
w/ Causal Reason + RerankContext-Description  & 0.65 \\ \hline
\end{tabular}
\end{table}

In the next set of experiments, we evaluate the performance of different LLMs at the reranking stage. As shown in \Cref{tab:rerank-methods-gold-llm} applying the w/ RerankKey prompt to GPT-4o increases the HitRate@1 score from 0.52 to 0.66, yielding an improvement of 14 percentage points. The w/ RerankContext prompt achieves comparable results for both GPT-4o and Claude 3.7. Given these similar outcomes and considering computational efficiency, we select GPT-4o as the reranker for all subsequent experiments and exclude the other LLMs from further analysis.

To assess the impact of reasoning-based prompting, we compare regular and causal reasoning strategies in the reranking process. \Cref{tab:rerank-methods} presents the top-1 accuracy results using GPT-4o, the best-performing reranker from \Cref{tab:rerank-methods-gold-llm}. The results show that incorporating reasoning (i.e. models starting with ``w/ Reason'') leads to a decrease of 1 percentage point in HitRate@1 performance compared to the baseline. However, both w/ Causal Reason + RerankKey prompt and w/ Causal Reason + RerankContext prompt improve the scores by 2 percentage points, reaching the highest observed HitRate@1 of 0.68. This demonstrates that explicitly prompting the LLM to consider cause-and-effect relationships between the retrieved skill and the job context enhances semantic relevance beyond surface similarity.

On the other hand, augmenting prompts with descriptions of ESCO labels or requiring reasoning over these descriptions led to a decrease in retrieval accuracy. These results suggest that the inclusion of causal reasoning allows the system to better account for practical relevance, rather than relying purely on linguistic similarity. Based on these findings, all subsequent end-to-end evaluations were conducted using causal reasoning prompts to ensure optimal skill linking accuracy.

As explained in \Cref{sec:end-to-end-eval}, we conduct a job-post-level skill linking evaluation of our end-to-end system using the best-performing skill identification results reported in Table 3 (i.e., BertTürk and Claude Sonnet 3.7 for the CoNLL-F1 score, and Gemma 3 for the MUC Partial score). As previously noted, due to computational constraints, we discontinue the use of LLMs other than GPT-4o for reranking and limit subsequent evaluations to Claude Sonnet 3.7 for skill identification. For reference, the average cost of each strategy for skill identification and reranking is provided in Appendix, \Cref{tab:api-cost}. Each skill identification prompt (such as zero-shot or few-shot) is executed only once. In contrast, each reranking strategy is run separately for each evaluation run. Since every configuration is evaluated across three runs, and all identification and reranking combinations are considered, the total cost of the complete evaluation amounts to approximately \$967. Consequently, we proceed with the remaining experiments using Claude Sonnet 3.7 for skill identification and GPT-4o for reranking.\footnote{We also ran the end-to-end evaluation with the other front-runner model, Gemma 3, which achieved a slightly lower score compared to Claude Sonnet 3.7; its results (with standard error) are reported in Appendix, in \Cref{tab:skillextractionresultswithstandarderrorsgemma} Gemma 3 was evaluated locally, allowing us to include it despite budgetary constraints. Additionally, we report GPT-4o results in Appendix, \Cref{tab:skillextractionresultswithgpt4o} to explore the performance when the same LLM is used for both skill identification and reranking; note that this configuration was run only once due to its higher cost.} \Cref{tab:skillextractionresultswithstandarderrors} presents the results of the end-to-end skill extraction pipeline. It also provides three different prompting strategies with causal reasoning (Columns 3–5), introduced in \Cref{section:llmRerank}, for reranking the retrieved candidates. The second column of the table provides our skill identification methods, whereas the third column (w/o Rerank) provides the performances of extracting the correct ESCO IDs with a baseline which does not use any reranking layer. The table provides its results under two main blocks for Fuzzy Matching and Embedding Similarity as our two skill retrieval strategies. This evaluation pipeline serves as an ablation study, demonstrating the impact of different components on the final skill extraction performance by comparing combinations of embedding similarity and fuzzy matching, reranking or no reranking, and reasoning or no reasoning.  Each configuration was run three times, and standard errors were included in the table to reflect the variability across runs.

The best-performing configuration in \Cref{tab:skillextractionresultswithstandarderrors} is Claude Sonnet 3.7 with dynamic ten-shot prompting for skill identification, embedding similarity for retrieval, and GPT-4o w/ RerankContext- Desc. Prompt with causal reasoning for reranking. This configuration achieved an average score of 0.56. While Claude Sonnet 3.7 with dynamic ten-shot prompting achieved the highest CoNLL-F1 score (0.56) among the LLMs, this score remains lower than the fully supervised BERT baseline.  When we investigate the situation, the manual annotations in our dataset tend to be short and keyword-based (e.g., ``machine learning'', ``deep learning'') rather than more descriptive phrases (e.g., ``knowledge of machine learning'' or ``proficient in deep learning''). On the other hand, Claude Sonnet 3.7 tends to generate longer skill spans that resemble the descriptive format of ESCO skills, which explains the lower performance on the CoNLL identification scores but the higher performance on the end-to-end evaluation. This suggests that Claude Sonnet 3.7's generated spans are more comprehensive and better matched to the ESCO skill base, despite the trade-off with the identification task. These results support our assumption about the usefulness of the MUC Partial score evaluations in skill identification.

\begin{table}
\centering
\caption{Results for End-to-End Skill Extraction: Identification with Claude Sonnet 3.7,  Reranking by GPT-4o w/ Causal Reasoning.}
\label{tab:skillextractionresultswithstandarderrors}
\begin{tabular}{c|l|cccc}

\multirow{2}[0]{*}{\mbox{
\begin{tabular}[c]{@{}c@{}}\textbf{\textbf{Retrieval }}\\\textbf{\textbf{Methods}}\end{tabular}
} }

 & \multirow{2}[0]{*}{\mbox{
\begin{tabular}[c]{@{}c@{}}\textbf{\textbf{Identification }}\\\textbf{\textbf{Methods}}\end{tabular}
} } &  \multicolumn{4}{c}{\textbf{Reranking Methods}} \\ 
 &  &  \begin{tabular}[c]{@{}c@{}}\textbf{\textbf{w/o }}\\\textbf{\textbf{~Rerank}}\end{tabular}
 & \begin{tabular}[c]{@{}c@{}}\textbf{\textbf{w/ RerankKey}}\\\textbf{\textbf{~Prompt}}\end{tabular} & \begin{tabular}[c]{@{}c@{}}\textbf{\textbf{w/ RerankContext }}\\\textbf{\textbf{Prompt}}\end{tabular} &
\begin{tabular}[c]{@{}c@{}}\textbf{\textbf{w/ RerankContext-}}\\\textbf{\textbf{~Desc. Prompt}}\end{tabular}
\\ \hhline{------}
\multirow{8}[15]{*}{\rotcell {\mbox{
\begin{tabular}[c]{@{}c@{}}\textbf{\textbf{Fuzzy }}\\\textbf{\textbf{Matching}}\end{tabular}}}} 
& SSL via Bert~ & $0.17\pm_{0.000}$ & $0.29\pm_{0.000}$ & $0.30\pm_{0.000}$ & $0.29\pm_{0.000}$ \\
 & Zero-Shot & $0.14\pm_{0.000}$ & $0.26\pm_{0.003}$ & $0.27\pm_{0.003}$ & $0.26\pm_{0.005}$   \\
 & Static Two-Shot & $0.16\pm_{0.000}$ & $0.28\pm_{0.005}$ & $0.28\pm_{0.003}$ & $0.28\pm_{0.003}$   \\
 & Static Six-Shot & $0.15\pm_{0.003}$ & $0.24\pm_{0.005}$ & $0.24\pm_{0.000}$ & $0.25\pm_{0.003}$   \\
 & Static Ten-Shot& $0.14\pm_{0.003}$ & $0.25\pm_{0.003}$ & $0.25\pm_{0.005}$ & $0.25\pm_{0.000}$   \\
 & Dynamic Two-Shot & $0.20\pm_{0.003}$ & $0.30\pm_{0.003}$ & $0.30\pm_{0.003}$ & $0.30\pm_{0.003}$   \\
 & Dynamic Six-Shot & $0.21\pm_{0.000}$ & $0.34\pm_{0.000}$ & $0.34\pm_{0.000}$ & $0.33\pm_{0.000}$   \\
 & Dynamic Ten-Shot & $0.22\pm_{0.000}$ & $0.34\pm_{0.003}$ & $0.34\pm_{0.000}$ & $0.34\pm_{0.003}$    \\ \hhline{------}
\multirow{8}[18]{*}{\rotcell {\mbox{
\begin{tabular}[c]{@{}c@{}}\textbf{\textbf{Embedding }}\\\textbf{\textbf{Similarity}}\end{tabular}}}} 
& SSL via Bert~ & $0.19\pm_{0.000}$ & $0.35\pm_{0.000}$ & $0.36\pm_{0.000}$ & $0.37\pm_{0.000}$   \\
 & Zero-Shot & $0.39\pm_{0.003}$ & $0.50\pm_{0.005}$ & $0.51\pm_{0.003}$ & $0.53\pm_{0.007}$   \\
 & Static Two-Shot & $0.40\pm_{0.003}$ & $0.54\pm_{0.005}$ & $0.54\pm_{0.005}$ & $0.55\pm_{0.003}$   \\
 & Static Six-Shot & $0.39\pm_{0.005}$ & $0.52\pm_{0.012}$ & $0.53\pm_{0.005}$ & $0.53\pm_{0.007}$   \\
 & Static Ten-Shot & $0.38\pm_{0.003}$ & $0.50\pm_{0.007}$ & $0.52\pm_{0.003}$ & $0.52\pm_{0.005}$   \\
 & Dynamic Two-Shot & $0.40\pm_{0.003}$ & $0.54\pm_{0.005}$ & $0.54\pm_{0.005}$ & $0.54\pm_{0.005}$   \\
 &  Dynamic Six-Shot & $0.40\pm_{0.003}$ & $0.52\pm_{0.003}$ & $0.53\pm_{0.008}$ & $0.55\pm_{0.000}$   \\
 & Dynamic Ten-Shot & $0.41\pm_{0.000}$ & $0.55\pm_{0.007}$ & $0.55\pm_{0.003}$ & $0.56\pm_{0.003}$   \\ \\ \hhline{------}
\end{tabular}
\end{table}

The complexity of skill extraction extends beyond simple keyword matching, as it requires a deeper semantic understanding for mapping to ESCO. For example, ``Word'' is not explicitly listed as a skill in ESCO but may be subsumed under terms like ``spreadsheet software'', requiring more nuanced semantic linking. Likewise, when linking terms like ``creative'', the challenge is understanding that this can be mapped to a broader description like ``think creatively'', demonstrating the need for semantic expertise in skill extraction. This complexity explains why embedding similarity outperforms fuzzy matching during the skill linking phase. Embedding-based methods capture the semantic relationships between skills more effectively, allowing for better alignment with the broader and more contextually nuanced ESCO skill categories. The improvement seen (columns \#4-6 vs column \#3) in the reranking process with GPT further highlights the importance of this approach. By leveraging GPT's semantic understanding, the model is able to refine skill matches based on context and meaning, rather than relying solely on surface-level keyword overlaps. This enhanced alignment with the ESCO database is the primary source of improvement in the overall skill extraction and linking performance.

Although the results are not strictly comparable with those reported in the literature due to different evaluation strategies applied to different approaches (e.g., end-to-end vs linking alone) and at different levels (e.g., mention vs sentence-level), we observe that our results are on par with ongoing studies using ESCO skill base.  For example, \citet{zhang-etal-2024-entitylink} focuses solely on the linking stage and reports a mention-level linking accuracy of 0.23 @k=1 for English, while \citet{llmranking} report a sentence-level RP@1 score of 0.46 and \citet{decorte2023extreme} report 0.55 RP@5, both in the technology domain for English.

\begin{table}[htb]
\caption{Key Findings from Expert Evaluation}
\label{tab:keyfindings}
\begin{tabular}{lp{12cm}}
\textbf{Key Feedback Type}          & \textbf{Summary of Expert Reviews}                    \\\hline
Essential Skill review     & The system accurately extracted numerous essential skills such as ``perform brand analysis’’, ``use microsoft office’’, `` manage cash flow’’, and ``SQL’’.   \\\hline
Optional skill review      & The system extracted nice-to-have skills too which is important in the process of publishing job ads. For example, the experts noted that , ``follow work procedures’’, or ``assume responsibility’’ are skills, but in some cases can be competencies or tasks too . Their comment stated: generally we do not consider "assume responsibility as a skill, but can be a duty or responsibility that we expect from candidates or job seekers…." \\\hline
Skill Omission             & In several instances, the system ignored some critical tools and technologies brand names explicitly mentioned in the Turkish job advertisement. For example, it missed ``Power BI’’ from a marketing and reporting role (Job ID 3661718) and the ``Logo program’’ from a finance position (Job ID 3662366), both of which are essential requirements.   \\\hline
Ambiguity in skill context & Some skills should be presented with fuller contextual expressions (as complete sentences). For example ``product price” which corresponds to ``fiyat ürün’’ in Turkish can cause ambiguity due to differences in sentence structure. In Turkish, the more natural and commonly used form is “ürün fiyatını belirleme” or  “Ürün fiyatlandırması yapma”. Although this skill is listed as in ESCO, it has been noted that such phrasing adjustments help reduce ambiguity arising from structural differences between Turkish and English. \\
\hline
\end{tabular}
\end{table}

At the human evaluation phase, the qualitative analysis is performed on a total of 230 skill extractions from 23 distinct job IDs.  That is, each job advertisement yielded about 10 automatically extracted skills on average. Our human evaluation findins are as follows: The system is highly effective at identifying what are perceived as the most critical skills per job advertisement and successfully identifies a significant number of "nice-to-have" skills. Statistically, the extraction performance strength lies in identifying essential job requirements, which accounted for 169 (73\%) of all extractions and categorized as Red (\Cref{sec:humeval}) which means essential skills correctly extracted. In addition, the system correctly identified 39 (approximately 17\%) optional or ``nice-to-have'' skills, labeled as Blue. The main area for improvement is highlighted by the 22 (10\%) extractions categorized as Gray, which represent omissions, ambiguities, or irrelevant terms. \Cref{tab:keyfindings}  summarizes these findings with illustrative examples from the experts' reviews and \Cref{appendix:humeval} in \Cref{sec:additionalresults} provides details of colour-coded analysis. This analysis based on human reviews provides a qualitative baseline for future model enhancements, reminding specific focus on reducing the omissions of locally relevant skills.

\section{Error Analysis}
\label{sec:error-analysis}

To better understand the limitations of our system, we conducted an error analysis on the test set, using the best performing configuration, which contains 340 manually labeled skill spans.

As part of this analysis, we first examined the output of the skill identification stage. Manual inspection revealed that, beyond unrelated extraction errors, the LLM-based models exhibited a consistent tendency to incorrectly label job position names as skills. This appears to stem from the fact that such positions often imply certain skill-sets, prompting the model to extract component terms as skills. For instance, the position title ``pazarlama ve ürün uzmanı'' (marketing and product specialist) was mistakenly split into two separate skill spans, ``pazarlama'' (marketing) and ``ürün'' (product), despite neither being labeled as skills in the gold standard annotations. These observations point to a challenge in distinguishing between explicit skill mentions and broader occupational references that merely imply skill usage.

Afterwards, this analysis focused specifically on the skill linking stage, where the majority of errors stemmed from semantic mismatches between extracted spans and entries in the ESCO skill base. Common issues included linking to overly generic concepts, context-inappropriate matches, or semantically irrelevant results due to surface-level similarity in embedding-based retrieval.

\begin{figure}[ht]
    \centering
    \includegraphics[width=\linewidth]{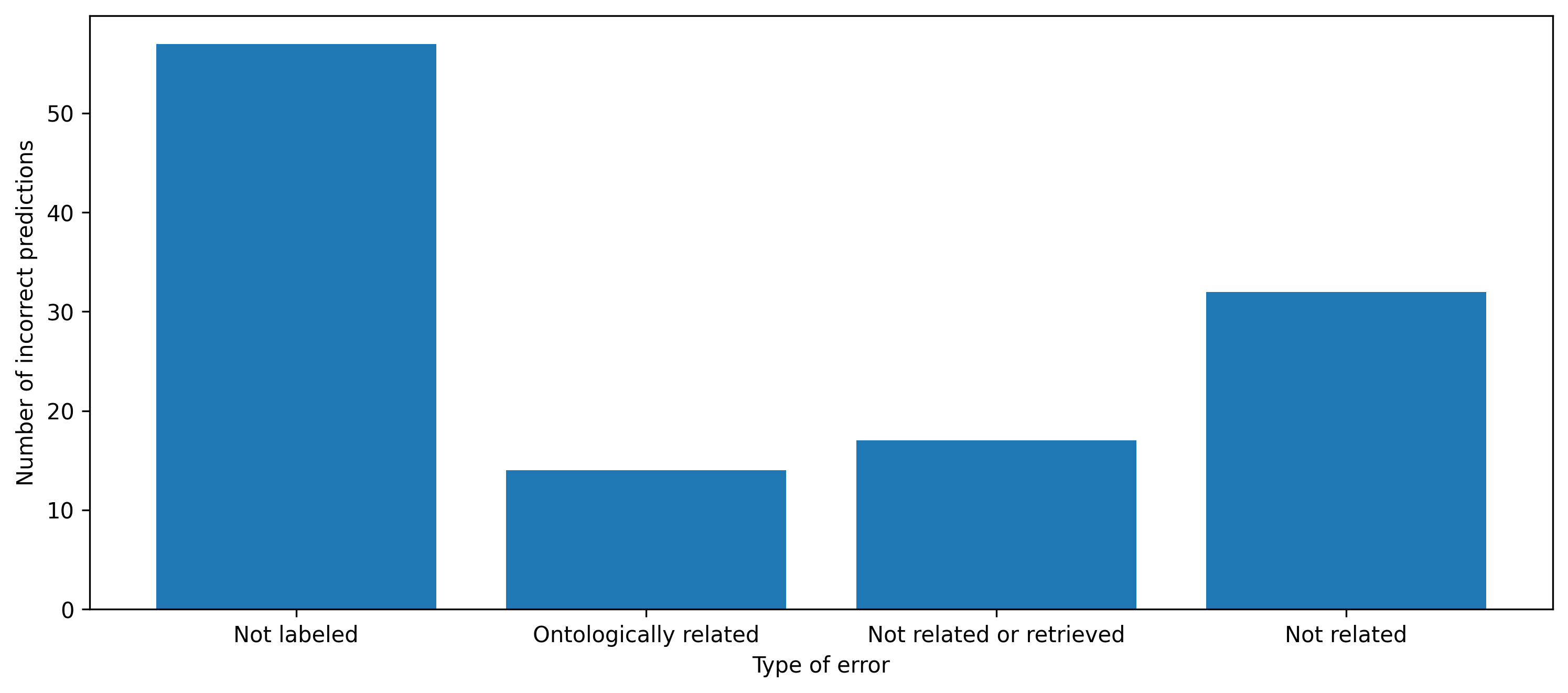}
    \caption{Distribution of error types in the LLM reranking step, categorized by error type.}
    \label{fig:error_distribution}
\end{figure}

We concentrated our analysis on the LLM reranking step within this stage. In many cases, the model linked a span to a skill that was not the gold-standard label but was still ontologically related, for example, a broader or narrower skill in the ESCO hierarchy, or a sibling skill sharing a common parent. To better characterize the nature of these errors, we categorized them into four types: (i) ``Not labeled'': Spans with no equivalent in the skill base, (ii) ``Ontologically related'': Predictions that are ontologically related but not exact matches, (iii) ``Not related or retrieved'': Cases where the correct skill was not retrieved and the predicted one was unrelated, and (iv) ``Not related'': entirely unrelated predictions. \Cref{fig:error_distribution} presents the distribution of these error categories. Of the $120$ total errors identified, $14$ ($11.67\%$) were classified as ``Ontologically Related''. For instance, majority of the errors in the ``Ontologically Related'' category involve the skill ``raporlamak (reporting/to report)''. Annotators consistently link this mention to the ESCO skill ``işle ilgili raporlar yazmak (write work-related reports)'', whereas our system links it to another ESCO skill within the same group, ``durum raporları yazmak (write situation reports)''. Similarly, the extracted span ``fiyat çalışmalarını'' was intended to be linked to the ESCO concept ``fiyat ürün (price products)'' but was instead mapped to ``fiyat önerilerinde bulunmak (make price recommendations)''. Although this match is technically incorrect, both skills belong to the same broader ESCO concept — ``determining values of goods or services''. These examples illustrate that ESCO’s hierarchical organization, while semantically coherent, introduces ambiguous boundaries between related nodes that can mislead embedding-based linking systems relying on cosine similarity alone.

Such errors reveal a limitation of current evaluation metrics, which penalize semantically related predictions as fully incorrect. Yet, these relations reflect meaningful conceptual proximity that a taxonomy-aware system could leverage. Incorporating hierarchical distance into the reranking process or multi-level classification objectives could allow the model to better respect ESCO’s structure and reduce these borderline errors. Likewise, developing hierarchy-sensitive evaluation metrics that assign partial credit based on ontology distance would provide a more realistic measure of system performance.

\section{Conclusion}
\label{sec:conclusion}

This article sought ways to leverage LLMs for Turkish skill extraction, addressing key challenges such as the lack of a comprehensive Turkish skill taxonomy, the absence of a skill extraction dataset and the morphologically rich nature of the language. We introduced the first Turkish skill extraction dataset containing labeled skill spans  on Turkish job postings. To enable evaluation of skill linking performance, we established references to the ESCO skill taxonomy only for the test set. A full translation of the taxonomy was beyond the scope of this work. Our findings highlighted the critical role of LLMs in overcoming these challenges, particularly through context-aware skill identification and linking. While our baseline supervised sequence labeling model outperforms both Claude Sonnet 3.7 and Gemma 3 in skill span identification, these decoder-only LLMs excel in the end-to-end pipeline, especially in mapping skill spans to the ESCO skill base. These results suggest that LLMs, when applied with effective prompting and retrieval techniques, can significantly improve skill extraction in low-resource settings, removing the need for data annotation to train supervised models.

Skill-based recruitment is rapidly gaining importance, shifting the focus from traditional job posting analysis to the identification of specific skills required for job roles. This transition is especially relevant in the context of AI-driven job matching solutions. By automating skill extraction from job postings, the methods explored in this article lay the groundwork for advanced applications such as AI-powered search and dynamic text-based recommendations, which can significantly improve the job matching process by ensuring a more precise alignment between job posts and the required skills. 

The complexity of the skill linking stage, driven by the high number of skill descriptions in ESCO, ambiguities in skill terms, and discrepancies in the level of detail between job posts and ESCO categories, appears to hinder high performance. As observed from the current literature, including our results, there is still room for improvement in achieving end-to-end high-performance skill extraction.

\section{Limitations \& Future Work}
\label{sec:limitations-future-work}

While this study offers the first Turkish dataset for skill extraction and presents a comparative evaluation of BERT-based and LLM-based models, it also comes with several limitations that warrant discussion.

First, although our dataset includes high-quality manual annotations, the annotation process was constrained by time and labor, resulting in a relatively limited number of labeled spans (i.e., 327 job postings). Nevertheless, as the first dataset of its kind for Turkish, it directly addresses our central research question of how skill extraction can be effectively performed for this low-resource language. While the dataset is comparable in size to those used in related work (\Cref{tab:datasets}), its source is a single employment platform, which may restrict representativeness across industries and job types (\Cref{fig:areadistribution}). Since preparing annotated training data is a demanding and costly process, the study proposes a method based on LLMs that can operate effectively even in low-resource settings. In this approach, labeled data samples in the training split of the dataset are used only in the skill identification stage. Therefore, although the sectoral distribution and representativeness of the test data are not perfectly balanced, the proposed method is considered adaptable to different sectors with minimal data preparation effort.

Second, our experiments showed that the BERT-based model outperformed the LLM-based approaches in skill identification, likely because the BERT model was fine-tuned directly on our labeled data while LLMs could not be fine-tuned due to resource constraints and lacked domain adaptation to recruitment or HR content. As a result, LLMs struggled to capture the specific vocabulary and structure of Turkish job postings. In addition, our evaluation required multiple LLM calls per configuration for reranking strategies and multi-shot prompting, incurring roughly \$967 in compute expenses, which limited our ability to repeat runs for statistical significance. Although few-shot prompting provided some task awareness, parameter-efficient fine-tuning methods such as LoRA or adapters, runnable locally without third-party API fees, could offer more robust alignment; future work should explore these strategies as practical alternatives to full model fine-tuning.

Furthermore, while our evaluation used standard span-based metrics (e.g., CoNLL-F1) to assess identification quality, these metrics may not fully capture the downstream usefulness of extracted skills in tasks such as linking or recommendation. Our results showed that Claude Sonnet 3.7, despite lower identification scores, performed competitively in end-to-end evaluations. This suggests that future studies should consider both span accuracy and semantic matching quality when designing evaluation frameworks.

Our error analysis further revealed that a non-trivial portion of linking errors resulted from the system predicting ontologically related skills (e.g., broader or narrower terms) rather than exact matches. However, current evaluation metrics penalize these semantically reasonable predictions the same as entirely unrelated ones. This limitation points to two complementary directions for future work. First, future work could focus on developing taxonomy-aware reranking methods that explicitly utilize hierarchical relationships between skills to improve semantic alignment and reduce such linking errors. Second, the design of evaluation metrics should be revisited to account for partial credit in cases where predictions are semantically close to gold labels. For instance, metrics that incorporate skill hierarchy distance or taxonomy-informed similarity could provide a more informative assessment of real-world performance, particularly in applications where capturing skill intent is more important than exact label matching.

In addition, the hierarchical structure of the ESCO taxonomy itself can be systematically leveraged in model design. Many of the observed errors occurred when the predicted skill belonged to a broader or narrower node of the correct concept, suggesting that current linking approaches lack hierarchical awareness. Future work could therefore explore hierarchical loss functions or multi-level classification frameworks that model these parent–child relations directly, allowing the system to penalize conceptually distant predictions more strongly than nearby ones. Such methods, combined with taxonomy-aware reranking, could transform the hierarchical nature of ESCO from a source of ambiguity into a structural advantage for improving semantic alignment.

As future work, we suggest expanding the dataset by incorporating job postings from multiple employment platforms, rather than relying on a single source. This would allow us to capture a broader range of industries and job types, thereby increasing representativeness of the Turkish labor market. In particular, care should be taken to balance the area distributions so that domains that are currently underrepresented (e.g., medicine and health) are more proportionally reflected compared to domains with higher representation (e.g., finance and sales/marketing), as illustrated in \Cref{fig:areadistribution}. Exploring cross-lingual or multilingual LLMs, domain-specific continued pretraining, and ethical considerations around dataset representativeness and potential bias are also promising directions. Finally, fine-tuning or adapting LLM components used in both identification and linking tasks could substantially improve system accuracy, robustness, and real-world applicability. To efficiently expand the training dataset and improve knowledge base for the Turkish labor market, an LLM-based data augmentation for extending the difference of the Turkish labor market could be employed. This process may provide proficiency in local tools, technologies, terminologies or  particular soft skills mentioned in Turkish job advertisements.

\section*{Practical Implications}
\label{sec:practical-implications}

The methods and dataset introduced in this study have immediate applications across the Turkish labor market ecosystem. Recruitment platforms can integrate our automated skill identification and linking pipeline to process unstructured job postings at scale, automatically tagging and standardizing skills, which both reduces recruiter workload and improves the precision of candidate–job matching and job recommendation engines.

Policymakers will find structured, language-specific skill data invaluable for curriculum design and upskilling policies. The ability to see which skills are most frequently requested, down to their exact phrasing in Turkish, enables more responsive education and certification standards that align with market realities.

Because our pipeline uses large language models, it can adapt to evolving terminology and domain-specific jargon found in job advertisements. This adaptability underpins advanced AI-powered applications, such as personalized career-path guidance, dynamic talent sourcing, and automated skill gap analysis, ensuring that new or niche skills are captured as they enter the marketplace.

Finally, by addressing the linguistic and resource constraints unique to Turkish, our work establishes a blueprint for extending skill extraction to other low-resource languages and specialized sectors. The methodologies and dataset we provide can be adapted and expanded, fostering global adoption of more inclusive, accurate, and scalable skill extraction technologies.

\section*{Data availability}
The data is sourced from a proprietary platform and is publicly released alongside this paper, with the access link provided.

\section*{Acknowledgments}
The authors would like to express their sincere gratitude to Kariyer.net for supporting this research through employment of the authors and for providing resources for the collection, analysis, and interpretation of data. We also thank WeHire for their valuable support in conducting expert human evaluations. In addition, we would like to offer our special thanks to all of our reviewers for their insightful comments, which have significantly improved the final version of this article. Finally, we thank our colleague Uygar Takazoğlu for his valuable discussions and support.

\bibliographystyle{unsrtnat}
\bibliography{refs}

@book{beaumont1993,
  title={Human resource management: {K}ey concepts and skills},
  author={Beaumont, Phil B},
  year={1993},
  publisher={Sage}
}

@article{ipmref,
title = {Skill requirements in job advertisements: A comparison of skill-categorization methods based on wage regressions},
journal = {Information Processing \& Management},
volume = {60},
number = {2},
pages = {103185},
year = {2023},
issn = {0306-4573},
doi = {https://doi.org/10.1016/j.ipm.2022.103185},
url = {https://www.sciencedirect.com/science/article/pii/S0306457322002862},
author = {Ziqiao Ao and Gergely Horváth and Chunyuan Sheng and Yifan Song and Yutong Sun}
}

@book{armstrong2023,
  title={Armstrong's {H}andbook of {H}uman {R}esource {M}anagement {P}ractice: {A} {G}uide to the {T}heory and {P}ractice of {P}eople management},
  author={Armstrong, Michael and Taylor, Stephen},
  year={2023},
  publisher={Kogan Page Publishers}
}

@article{Skill-currency-OECD,
  author  = {Glenda Quintini},
  title   = {Skills at Work: How Skills and their Use Matter in the Labour Market},
  journal = {OECD Social, Employment and Migration Working Papers},
  year    = {2014},
  url     = {https://www.oecd-ilibrary.org/content/paper/5jz44fdfjm7j-en},
  doi     = {10.1787/5jz44fdfjm7j-en}
}

@inproceedings{ranaldi-etal-2024-empowering,
    title = "Empowering cross-lingual abilities of instruction-tuned large language models by translation-following demonstrations",
    author = "Ranaldi, Leonardo  and
      Pucci, Giulia  and
      Freitas, Andre",
    editor = "Ku, Lun-Wei  and
      Martins, Andre  and
      Srikumar, Vivek",
    booktitle = "Findings of the Association for Computational Linguistics: ACL 2024",
    month = aug,
    year = "2024",
    address = "Bangkok, Thailand",
    publisher = "Association for Computational Linguistics",
    url = "https://aclanthology.org/2024.findings-acl.473/",
    doi = "10.18653/v1/2024.findings-acl.473",
    pages = "7961--7973",
    
}

@article{
kiciman2024causal,
title={Causal Reasoning and Large Language Models: Opening a New Frontier for Causality},
author={Emre Kiciman and Robert Ness and Amit Sharma and Chenhao Tan},
journal={Transactions on Machine Learning Research},
issn={2835-8856},
year={2024},
url={https://openreview.net/forum?id=mqoxLkX210},
note={Featured Certification}
}

@incollection{Ramshaw1999,
  author    = {Ramshaw, L. A. and Marcus, M. P.},
  title     = {Text Chunking Using Transformation-Based Learning},
  booktitle = {Natural Language Processing Using Very Large Corpora},
  editor    = {Armstrong, Susan and Church, Kenneth and Isabelle, Pierre and Manzi, Sandra and Tzoukermann, Evelyne and Yarowsky, David},
  year      = {1999},
  publisher = {Springer Netherlands},
  address   = {Dordrecht},
  pages     = {157--176},
  isbn      = {978-94-017-2390-9},
  doi       = {10.1007/978-94-017-2390-9_10},
  url       = {https://doi.org/10.1007/978-94-017-2390-9_10}
}

@inproceedings{conll,
  title={Introduction to the {CoNLL}-2003 Shared Task: Language-Independent Named Entity Recognition},
  author={Sang, Erik Tjong Kim and De Meulder, Fien},
  booktitle={Proceedings of the Seventh Conference on Natural Language Learning at HLT-NAACL 2003},
  pages={142--147},
  year={2003}
}

@inproceedings{zhang-etal-2024-entitylink,
    title = "Entity {L}inking in the {J}ob {M}arket {D}omain",
    author = "Zhang, Mike  and
      van der Goot, Rob  and
      Plank, Barbara",
    editor = "Graham, Yvette  and
      Purver, Matthew",
    booktitle = "Findings of the Association for Computational Linguistics: EACL 2024",
    month = mar,
    year = "2024",
    address = "St. Julian{'}s, Malta",
    publisher = "Association for Computational Linguistics",
    url = "https://aclanthology.org/2024.findings-eacl.28",
    pages = "410--419",
}

@misc{VQA-2019,
  author = {{Vocational Qualifications Authority of Türkiye}},
  title = {Implementation of {T}urkish Qualifications System and Framework Operation ({TUYEP})},
  note = {Action II. - Education and Training, Activity II.III. - Strengthening National Qualifications System and Promoting LLL \& Adaptability. Operation Number: TREESP2.3.TUYEP. Implementation Period: Operation: 25.11.2019--31.08.2025, Service component: 25.11.2019--29.02.2024, Grant component: 01.11.2020--31.03.2024, Direct Grant component: 01.09.2021--31.03.2024},
  year = {2019},
  url = {https://www.vqa.gov.tr}
}

@inproceedings{reranker,
   title={Large Language Model Is Not a Good Few-shot Information Extractor, but a Good Reranker for Hard Samples!},
   url={http://dx.doi.org/10.18653/v1/2023.findings-emnlp.710},
   DOI={10.18653/v1/2023.findings-emnlp.710},
   booktitle={Findings of the Association for Computational Linguistics: EMNLP 2023},
   publisher={Association for Computational Linguistics},
   author={Ma, Yubo and Cao, Yixin and Hong, Yong and Sun, Aixin},
   year={2023} }

@inproceedings{muc,
    title = "{MUC}-5 Evaluation Metrics",
    author = "Chinchor, Nancy  and
      Sundheim, Beth",
    booktitle = "Fifth Message Understanding Conference ({MUC}-5): Proceedings of a Conference Held in Baltimore, {M}aryland, August 25-27, 1993",
    year = "1993",
    url = "https://aclanthology.org/M93-1007",
}

@inproceedings{partial,
    title = "{S}em{E}val-2013 Task 9 : Extraction of Drug-Drug Interactions from Biomedical Texts ({DDIE}xtraction 2013)",
    author = "Segura-Bedmar, Isabel  and
      Mart{\'\i}nez, Paloma  and
      Herrero-Zazo, Mar{\'\i}a",
    editor = "Manandhar, Suresh  and
      Yuret, Deniz",
    booktitle = "Second Joint Conference on Lexical and Computational Semantics (*{SEM}), Volume 2: Proceedings of the Seventh International Workshop on Semantic Evaluation ({S}em{E}val 2013)",
    month = jun,
    year = "2013",
    address = "Atlanta, Georgia, USA",
    publisher = "Association for Computational Linguistics",
    url = "https://aclanthology.org/S13-2056",
    pages = "341--350",
}

@inproceedings{nguyen-etal-2024-rethinking,
    title = "Rethinking Skill Extraction in the Job Market Domain using Large Language Models",
    author = "Nguyen, Khanh  and
      Zhang, Mike  and
      Montariol, Syrielle  and
      Bosselut, Antoine",
    editor = "Hruschka, Estevam  and
      Lake, Thom  and
      Otani, Naoki  and
      Mitchell, Tom",
    booktitle = "Proceedings of the First Workshop on Natural Language Processing for Human Resources (NLP4HR 2024)",
    month = mar,
    year = "2024",
    address = "St. Julian{'}s, Malta",
    publisher = "Association for Computational Linguistics",
    url = "https://aclanthology.org/2024.nlp4hr-1.3",
    pages = "27--42",
    
}

@book{isco,
  title={International Standard Classification of Occupations 2008 (ISCO-08): Structure, group definitions and correspondence tables},
  author={International Labour Office},
  year={2012},
  publisher={International Labour Office}
}

@mastersthesis{aslim2023,
    author       = "Gülşah Kargin Aslim",
    title        = "Assessment for {I}dentifying {S}kills {G}aps in {H}igh {P}erformance {C}omputing {R}elated {H}igher {E}ducation {P}rograms by {U}sing {NLP}",
    school       = "Middle East Technical University",
    year         = "2023",
    month        = apr,
    note         = "{Submitted by Gülşah KARGIN ASLIM in partial fulfillment of the requirements for the degree of Master of Science in Information Systems Department, Middle East Technical University}"
}

@misc{australia-2023,
    author = "{Jobs and {S}kills {A}ustralia}",
    title = "{Jobs and Skills Atlas Methodology}",
    year = "2023",
    month = nov,
    publisher = "{Australia Government}",
    url = "https://www.jobsandskills.gov.au/sites/default/files/2023-11/jobs_ad_skills_atlas_atlas_methodology_november_2023_0.pdf"
}

@inproceedings{llm-rethinking,
    title = "Rethinking {S}kill {E}xtraction in the {J}ob {M}arket {D}omain using {L}arge {L}anguage {M}odels",
    author = "Nguyen, Khanh  and
      Zhang, Mike  and
      Montariol, Syrielle  and
      Bosselut, Antoine",
    editor = "Hruschka, Estevam  and
      Lake, Thom  and
      Otani, Naoki  and
      Mitchell, Tom",
    booktitle = "Proceedings of the First Workshop on Natural Language Processing for Human Resources (NLP4HR 2024)",
    month = mar,
    year = "2024",
    address = "St. Julian{'}s, Malta",
    publisher = "Association for Computational Linguistics",
    url = "https://aclanthology.org/2024.nlp4hr-1.3",
    pages = "27--42",
}

@inproceedings{senger-etal-2024-deeplearning,
    title = "Deep {L}earning-based {C}omputational {J}ob {M}arket {A}nalysis: {A} {S}urvey on skill {E}xtraction and {C}lassification from {J}ob {P}ostings",
    author = "Senger, Elena  and
      Zhang, Mike  and
      Goot, Rob  and
      Plank, Barbara",
    editor = "Hruschka, Estevam  and
      Lake, Thom  and
      Otani, Naoki  and
      Mitchell, Tom",
    booktitle = "Proceedings of the First Workshop on Natural Language Processing for Human Resources (NLP4HR 2024)",
    month = mar,
    year = "2024",
    address = "St. Julian{'}s, Malta",
    publisher = "Association for Computational Linguistics",
    url = "https://aclanthology.org/2024.nlp4hr-1.1",
    pages = "1--15"
}

@inproceedings{nnose,
    title = "{NNOSE}: {N}earest {N}eighbor {O}ccupational {S}kill {E}xtraction",
    author = "Zhang, Mike  and
      Goot, Rob  and
      Kan, Min-Yen  and
      Plank, Barbara",
    editor = "Graham, Yvette  and
      Purver, Matthew",
    booktitle = "Proceedings of the 18th Conference of the European Chapter of the Association for Computational Linguistics (Volume 1: Long Papers)",
    month = mar,
    year = "2024",
    address = "St. Julian{'}s, Malta",
    publisher = "Association for Computational Linguistics",
    url = "https://aclanthology.org/2024.eacl-long.35",
    pages = "589--608"
}

@inproceedings{entitylinking,
    title = "Entity {L}inking in the {J}ob {M}arket {D}omain",
    author = "Zhang, Mike  and
      Goot, Rob  and
      Plank, Barbara",
    editor = "Graham, Yvette  and
      Purver, Matthew",
    booktitle = "Findings of the Association for Computational Linguistics: EACL 2024",
    month = mar,
    year = "2024",
    address = "St. Julian{'}s, Malta",
    publisher = "Association for Computational Linguistics",
    url = "https://aclanthology.org/2024.findings-eacl.28",
    pages = "410--419"
}

@inproceedings{llmranking,
  articleno    = {8},
  author       = {Benjamin Clavié and Guillaume Soulié},
  booktitle    = {Proceedings of the 3rd Workshop on Recommender Systems for Human Resources (RecSys in HR 2023)},
  editor       = {Kaya, Mesut and Bogers, Toine and Graus, David and Johnson, Chris and Decorte, Jens-Joris},
  isbn         = {9781450398565},
  issn         = {1613-0073},
  language     = {eng},
  location     = {Singapore, Singapore},
  pages        = {10},
  publisher    = {CEUR},
  title        = {Large Language Models as Batteries-Included Zero-Shot {ESCO} Skills Matchers},
  url          = {https://ceur-ws.org/Vol-3490/RecSysHR2023-paper_8.pdf},
  volume       = {3490},
  year         = {2023},
}

@inproceedings{gnehm2022,
    title = "{F}ine-{G}rained {E}xtraction and {C}lassification of {S}kill {R}equirements in {G}erman-{S}peaking {J}ob {A}ds",
    author = "Gnehm, Ann-Sophie and
      Bühlmann, Eva and
      Buchs, Helen and
      Clematide, Simon",
    booktitle = "Proceedings of the Fifth Workshop on Natural Language Processing and Computational Social Science (NLP+CSS)",
    month = dec,
    year = "2022",
    address = "Abu Dhabi",
    publisher = "Association for Computational Linguistics",
    pages = "14--24"
}

@book{onet2010,
  author    = {National Research Council},
  title     = {A Database for a Changing Economy: Review of the Occupational Information Network (O*NET)},
  year      = {2010},
  publisher = {The National Academies Press},
  address   = {Washington, DC},
  isbn      = {978-0-309-14769-9},
  doi       = {10.17226/12814},
  url       = {https://nap.nationalacademies.org/catalog/12814/a-database-for-a-changing-economy-review-of-the-occupational},
  note      = {Edited by Nancy T. Tippins and Margaret L. Hilton}
}

@ARTICLE{esco,
  author={le Vrang, Martin and Papantoniou, Agis and Pauwels, Erika and Fannes, Pieter and Vandensteen, Dominique and De Smedt, Johan},
  journal={Computer}, 
  title={{ESCO}: {B}oosting {J}ob {M}atching in {E}urope with {S}emantic {I}nteroperability}, 
  year={2014},
  volume={47},
  number={10},
  pages={57-64},
  doi={10.1109/MC.2014.283}}

@misc{lightcast-skill,
    author = "{Lightcast}",
    title = "{What are skills?}",
    year = "2024",
    url = "https://lightcast.io/what-are-skills",
    note = "Retrieved June 3, 2024"
}

@misc{esco-data,
    author = "{ESCO (V 1.2)}",
    title = "{What is {ESCO}?}",
    year = "2024",
    url = "https://esco.ec.europa.eu/en/about-esco/what-esco",
    note = "Retrieved June, 2024"
}

@inproceedings{green-etal-2022-development,
    title = "Development of a {B}enchmark {C}orpus to {S}upport {E}ntity {R}ecognition in {J}ob {D}escriptions",
    author = "Green, Thomas  and
      Maynard, Diana  and
      Lin, Chenghua",
    editor = "Calzolari, Nicoletta  and
      B{\'e}chet, Fr{\'e}d{\'e}ric  and
      Blache, Philippe  and
      Choukri, Khalid  and
      Cieri, Christopher  and
      Declerck, Thierry  and
      Goggi, Sara  and
      Isahara, Hitoshi  and
      Maegaard, Bente  and
      Mariani, Joseph  and
      Mazo, H{\'e}l{\`e}ne  and
      Odijk, Jan  and
      Piperidis, Stelios",
    booktitle = "Proceedings of the Thirteenth Language Resources and Evaluation Conference",
    month = jun,
    year = "2022",
    address = "Marseille, France",
    publisher = "European Language Resources Association",
    url = "https://aclanthology.org/2022.lrec-1.128",
    pages = "1201--1208"
}

@misc{fijo,
      title={"{FIJO}": a {F}rench {I}nsurance {S}oft {S}kill {D}etection {D}ataset}, 
      author={David Beauchemin and Julien Laumonier and Yvan Le Ster and Marouane Yassine},
      year={2022},
      eprint={2204.05208},
      archivePrefix={arXiv},
      primaryClass={cs.CL},
      url={https://arxiv.org/abs/2204.05208}, 
}

@inproceedings{zhang-etal-2022-kompetencer,
    title = "Kompetencer: {F}ine-grained {S}kill {C}lassification in {D}anish {J}ob {P}ostings via {D}istant {S}upervision and {T}ransfer {L}earning",
    author = "Zhang, Mike  and
      Jensen, Kristian N{\o}rgaard  and
      Plank, Barbara",
    editor = "Calzolari, Nicoletta  and
      B{\'e}chet, Fr{\'e}d{\'e}ric  and
      Blache, Philippe  and
      Choukri, Khalid  and
      Cieri, Christopher  and
      Declerck, Thierry  and
      Goggi, Sara  and
      Isahara, Hitoshi  and
      Maegaard, Bente  and
      Mariani, Joseph  and
      Mazo, H{\'e}l{\`e}ne  and
      Odijk, Jan  and
      Piperidis, Stelios",
    booktitle = "Proceedings of the Thirteenth Language Resources and Evaluation Conference",
    month = jun,
    year = "2022",
    address = "Marseille, France",
    publisher = "European Language Resources Association",
    url = "https://aclanthology.org/2022.lrec-1.46",
    pages = "436--447",
}

@misc{skillgpt,
      title={SkillGPT: a {REST}ful {API} service for skill extraction and standardization using a {L}arge {L}anguage {M}odel}, 
      author={Nan Li and Bo Kang and Tijl De Bie},
      year={2023},
      eprint={2304.11060},
      archivePrefix={arXiv},
      primaryClass={cs.CL},
      url={https://arxiv.org/abs/2304.11060} 
}

@InProceedings{sayfullina,
author="Sayfullina, Luiza
and Malmi, Eric
and Kannala, Juho",
editor="van der Aalst, Wil M. P.
and Batagelj, Vladimir
and Glava{\v{s}}, Goran
and Ignatov, Dmitry I.
and Khachay, Michael
and Kuznetsov, Sergei O.
and Koltsova, Olessia
and Lomazova, Irina A.
and Loukachevitch, Natalia
and Napoli, Amedeo
and Panchenko, Alexander
and Pardalos, Panos M.
and Pelillo, Marcello
and Savchenko, Andrey V.",
title="Learning {R}epresentations for {S}oft {S}kill {M}atching",
booktitle="Analysis of {I}mages, {S}ocial {N}etworks and {T}exts",
year="2018",
publisher="Springer International Publishing",
address="Cham",
pages="141--152",
isbn="978-3-030-11027-7"
}

@inproceedings{skillspan,
    title = "{S}kill{S}pan: Hard and Soft Skill Extraction from {E}nglish Job Postings",
    author = "Zhang, Mike  and
      Jensen, Kristian  and
      Sonniks, Sif  and
      Plank, Barbara",
    editor = "Carpuat, Marine  and
      de Marneffe, Marie-Catherine  and
      Meza Ruiz, Ivan Vladimir",
    booktitle = "Proceedings of the 2022 Conference of the North American Chapter of the Association for Computational Linguistics: Human Language Technologies",
    month = jul,
    year = "2022",
    address = "Seattle, United States",
    publisher = "Association for Computational Linguistics",
    url = "https://aclanthology.org/2022.naacl-main.366/",
    doi = "10.18653/v1/2022.naacl-main.366",
    pages = "4962--4984"}

@inproceedings{designnegative-decorte,
  articleno    = {4},
  author       = {Decorte, Jens-Joris and Van Hautte, Jeroen and Deleu, Johannes and Develder, Chris and Demeester, Thomas},
  booktitle    = {Proceedings of the 2nd Workshop on Recommender Systems for Human Resources (RecSys-in-HR 2022)},
  editor       = {Kaya, Mesut and Bogers, Toine and Graus, David and Mesbah, Sepideh and Johnson, Chris and Gutiérrez, Francisco},
  isbn         = {9781450398565},
  issn         = {1613-0073},
  language     = {eng},
  location     = {Seatle, USA},
  pages        = {7},
  publisher    = {CEUR},
  title        = {Design of negative sampling strategies for distantly supervised skill extraction},
  url          = {https://ceur-ws.org/Vol-3218/RecSysHR2022-paper_4.pdf},
  volume       = {3218},
  year         = {2022},
}

@inproceedings{bunt-mama,
  title     = {Towards Interoperable Annotation of Quantification},
  author    = {Bunt, Harry},
  booktitle = {Proceedings of the 13th Joint {ISO}-{ACL} Workshop on Interoperable Semantic Annotation ({ISA}-13)},
  year      = {2017},
  pages     = {48--54},
  url       = {https://aclanthology.org/W17-7409}
}

@Article{svm,
author={Cortes, Corinna
and Vapnik, Vladimir},
title={Support-vector networks},
journal={Machine Learning},
year={1995},
month={Sep},
day={01},
volume={20},
number={3},
pages={273-297},
issn={1573-0565},
doi={10.1007/BF00994018},
url={https://doi.org/10.1007/BF00994018}
}

@article{scikit-learn,
  title={Scikit-learn: Machine Learning in {P}ython},
  author={Pedregosa, F. and Varoquaux, G. and Gramfort, A. and Michel, V.
          and Thirion, B. and Grisel, O. and Blondel, M. and Prettenhofer, P.
          and Weiss, R. and Dubourg, V. and Vanderplas, J. and Passos, A. and
          Cournapeau, D. and Brucher, M. and Perrot, M. and Duchesnay, E.},
  journal={Journal of Machine Learning Research},
  volume={12},
  pages={2825--2830},
  year={2011}
}

@inproceedings{2024-jobskape,
    title = "{J}ob{S}kape: A Framework for Generating Synthetic Job Postings to Enhance Skill Matching",
    author = "Magron, Antoine  and
      Dai, Anna  and
      Zhang, Mike  and
      Montariol, Syrielle  and
      Bosselut, Antoine",
    editor = "Hruschka, Estevam  and
      Lake, Thom  and
      Otani, Naoki  and
      Mitchell, Tom",
    booktitle = "Proceedings of the First Workshop on Natural Language Processing for Human Resources (NLP4HR 2024)",
    month = mar,
    year = "2024",
    address = "St. Julian{'}s, Malta",
    publisher = "Association for Computational Linguistics",
    url = "https://aclanthology.org/2024.nlp4hr-1.4/",
    pages = "43--58"
}

@misc{rapidfuzz,
  author       = {Max Bachmann},
  title        = {rapidfuzz/RapidFuzz: Release 3.8.1},
  year         = {2024},
  month        = apr,
  howpublished = {\url{https://doi.org/10.5281/zenodo.10938887}},
  note         = {Version v3.8.1. Published on Zenodo},
  doi          = {10.5281/zenodo.10938887}
}

@misc{matkin2024comparative,
      title={Comparative Analysis of Encoder-Based {NER} and Large Language Models for Skill Extraction from Russian Job Vacancies}, 
      author={Nikita Matkin and Aleksei Smirnov and Mikhail Usanin and Egor Ivanov and Kirill Sobyanin and Sofiia Paklina and Petr Parshakov},
      year={2024},
      eprint={2407.19816},
      archivePrefix={arXiv},
      primaryClass={cs.CL},
      url={https://arxiv.org/abs/2407.19816}, 
}

@inproceedings{kavas-etal-2025-multilingual,
    title = "Multilingual Skill Extraction for Job Vacancy{--}Job Seeker Matching in Knowledge Graphs",
    author = "Kavas, Hamit  and
      Serra-Vidal, Marc  and
      Wanner, Leo",
    editor = "Gesese, Genet Asefa  and
      Sack, Harald  and
      Paulheim, Heiko  and
      Merono-Penuela, Albert  and
      Chen, Lihu",
    booktitle = "Proceedings of the Workshop on Generative AI and Knowledge Graphs (GenAIK)",
    month = jan,
    year = "2025",
    address = "Abu Dhabi, UAE",
    publisher = "International Committee on Computational Linguistics",
    url = "https://aclanthology.org/2025.genaik-1.15/",
    pages = "146--155"
}

@article{gavrilescu2025,
  title={Techniques for Transversal Skill Classification and Relevant Keyword Extraction from Job Advertisements},
  author={Gavrilescu, Marius and Leon, Florin and Minea, Alina-Adriana},
  journal={Information},
  volume={16},
  number={3},
  pages={167},
  year={2025},
  publisher={MDPI}
}

@article{careerbert2025,
  title = {CareerBERT: Matching Resumes to ESCO Jobs in a Shared Embedding Space for Generic Job Recommendations},
  author = {Rosenberger, Julian and Wolfrum, Lukas and Weinzierl, Sven and Kraus, Mathias and Zschech, Patrick},
  journal = {Expert Systems with Applications},
  volume = {275},
  pages = {127043},
  year = {2025},
  publisher = {ScienceDirect},
  doi = {10.1016/j.eswa.2025.127043}
}

@misc{2023llm4jobs,
      title={LLM4Jobs: Unsupervised occupation extraction and standardization leveraging Large Language Models}, 
      author={Nan Li and Bo Kang and Tijl De Bie},
      year={2023},
      eprint={2309.09708},
      archivePrefix={arXiv},
      primaryClass={cs.CL},
      url={https://arxiv.org/abs/2309.09708}, 
}

@article{gpt4o,
  title={Gpt-4o system card},
  author={Hurst, Aaron and Lerer, Adam and Goucher, Adam P and Perelman, Adam and Ramesh, Aditya and Clark, Aidan and Ostrow, AJ and Welihinda, Akila and Hayes, Alan and Radford, Alec and others},
  journal={arXiv preprint arXiv:2410.21276},
  year={2024}
}

@article{qwen3,
  title={Qwen3 technical report},
  author={Yang, An and Li, Anfeng and Yang, Baosong and Zhang, Beichen and Hui, Binyuan and Zheng, Bo and Yu, Bowen and Gao, Chang and Huang, Chengen and Lv, Chenxu and others},
  journal={arXiv preprint arXiv:2505.09388},
  year={2025}
}

@article{gemma3,
  title={Gemma 3 technical report},
  author={Team, Gemma and Kamath, Aishwarya and Ferret, Johan and Pathak, Shreya and Vieillard, Nino and Merhej, Ramona and Perrin, Sarah and Matejovicova, Tatiana and Ram{\'e}, Alexandre and Rivi{\`e}re, Morgane and others},
  journal={arXiv preprint arXiv:2503.19786},
  year={2025}
}

@article{eurobert,
  title={EuroBERT: Scaling Multilingual Encoders for European Languages},
  author={Boizard, Nicolas and Gisserot-Boukhlef, Hippolyte and Alves, Duarte M and Martins, Andr{\'e} and Hammal, Ayoub and Corro, Caio and Hudelot, C{\'e}line and Malherbe, Emmanuel and Malaboeuf, Etienne and Jourdan, Fanny and others},
  journal={arXiv preprint arXiv:2503.05500},
  year={2025}
}

@inproceedings{almoststochasticorder,
    title = "Deep Dominance - How to Properly Compare Deep Neural Models",
    author = "Dror, Rotem  and
      Shlomov, Segev  and
      Reichart, Roi",
    editor = "Korhonen, Anna  and
      Traum, David  and
      M{\`a}rquez, Llu{\'i}s",
    booktitle = "Proceedings of the 57th Annual Meeting of the Association for Computational Linguistics",
    month = jul,
    year = "2019",
    address = "Florence, Italy",
    publisher = "Association for Computational Linguistics",
    url = "https://aclanthology.org/P19-1266/",
    doi = "10.18653/v1/P19-1266",
    pages = "2773--2785"
}

@article{Decorte_2025,
   title={Efficient Text Encoders for Labor Market Analysis},
   volume={13},
   ISSN={2169-3536},
   url={http://dx.doi.org/10.1109/ACCESS.2025.3589147},
   DOI={10.1109/access.2025.3589147},
   journal={IEEE Access},
   publisher={Institute of Electrical and Electronics Engineers (IEEE)},
   author={Decorte, Jens-Joris and van Hautte, Jeroen and Develder, Chris and Demeester, Thomas},
   year={2025},
   pages={133596–133608} }

@misc{decorte2023extreme,
      title={Extreme Multi-Label Skill Extraction Training using Large Language Models}, 
      author={Jens-Joris Decorte and Severine Verlinden and Jeroen Van Hautte and Johannes Deleu and Chris Develder and Thomas Demeester},
      year={2023},
      eprint={2307.10778},
      archivePrefix={arXiv},
      primaryClass={cs.CL},
      url={https://arxiv.org/abs/2307.10778}, 
}

@misc{hu2021loralowrankadaptationlarge,
      title={LoRA: Low-Rank Adaptation of Large Language Models}, 
      author={Edward J. Hu and Yelong Shen and Phillip Wallis and Zeyuan Allen-Zhu and Yuanzhi Li and Shean Wang and Lu Wang and Weizhu Chen},
      year={2021},
      eprint={2106.09685},
      archivePrefix={arXiv},
      primaryClass={cs.CL},
      url={https://arxiv.org/abs/2106.09685}, 
}

@misc{dettmers2023qloraefficientfinetuningquantized,
      title={QLoRA: Efficient Finetuning of Quantized LLMs}, 
      author={Tim Dettmers and Artidoro Pagnoni and Ari Holtzman and Luke Zettlemoyer},
      year={2023},
      eprint={2305.14314},
      archivePrefix={arXiv},
      primaryClass={cs.LG},
      url={https://arxiv.org/abs/2305.14314}, 
}

\appendix
\section{Appendix}
\label{appendix}
This appendix provides the original prompts and their English translations used during skill identification and linking stages.

\subsection{Implementation Details}
\subsubsection{Transformer Fine-Tuning}
\label{sec:transformerfinetuning}
Both Turkish BERT and EuroBERT-210m models are fine-tuned using the Hugging Face Transformers library with the same hyperparameters:

\begin{itemize}
    \item Learning rate = $2 \cdot 10^{-5}$
    \item Weight decay = 0.01
    \item Epochs = 6
\end{itemize}

\subsubsection{Prompting Configuration}
\label{sec:promptconfig}
All LLMs are evaluated with a temperature of 0 to ensure deterministic outputs. For models supporting explicit reasoning or “thinking” modes (e.g., Claude and Gemini), these options are disabled to maintain comparability. For other models, only their default generation behavior is used without additional reasoning instructions.

\subsubsection{Dynamic Prompting}
\label{sec:dynamicprompt}
The kNN retriever uses k = 5 nearest examples, trained on the combined training + validation sets.

\subsubsection{Skill Retrieval}
\label{sec:appendixskillretrieval}
For fuzzy matching, we use Rapidfuzz with $token\_sort\_ratio$ as the scoring function. For embedding-based similarity, we employ the multilingual-e5-large sentence-transformers model\footnote{\url{https://huggingface.co/intfloat/multilingual-e5-large}}, and compute cosine similarity between extracted skill spans and translated ESCO entries.

\subsubsection{LLM Reranking}
\label{sec:appendixllmreranking}
Reranking steps reuse the same LLM set listed in \Cref{sec:experimentalsetup} for Skill Identification with temperature = $0$ for reproducibility.

\subsection{Skill Identification Results on Sales-Marketing and Finance Areas}
\label{section:appendixArea}

\begin{table}[htp]
\centering
\caption{Average Skill Identification Results on Sales-Marketing and Finance Areas with Claude Sonnet 3.7 }
\label{tab:appendixClaudeArea}
\begin{tabular}{c|lc||c|lc}
\multicolumn{1}{c|}{\textbf{Area}} &
\begin{tabular}[c]{@{}c@{}}\textbf{Prompt}\\\textbf{~Strategy}\end{tabular} & \begin{tabular}[c]{@{}c@{}}\textbf{MUC}\\\textbf{~Partial}\end{tabular} & \multicolumn{1}{c|}{\textbf{Area}} & \begin{tabular}[c]{@{}c@{}}\textbf{Prompt}\\\textbf{~Strategy}\end{tabular} & \begin{tabular}[c]{@{}c@{}}\textbf{MUC}\\\textbf{~Partial}\end{tabular}\\ \hline

\multirow{7}[20]{*}{\rotcell{\mbox{\textbf{Sales-Marketing}}}} 
& Zero-Shot& 0.63 & 
\multirow{7}[10]{*}{\rotcell{\mbox{\textbf{Finance}}}} 
& Zero-Shot& 0.66 \\
& Static Two-Shot  & 0.78 &
    & Static Two-Shot  &    0.81 \\
& Static Six-Shot  & 0.81 &
    & Static Six-Shot   & 0.77 \\
& Static Ten-Shot   & 0.78 &
    & Static Ten-Shot  &  0.79 \\
& Dynamic Two-Shot & 0.73 &          
    & Dynamic Two-Shot & 0.77 \\
& Dynamic Six-Shot  & 0.75 &          
    & Dynamic Six-Shot & 0.76 \\
& Dynamic Ten-Shot &  0.78 &          
    & Dynamic Ten-Shot &  0.79 \\ \hline
\end{tabular}
\end{table}

\begin{table}[htp]
\centering
\caption{Average Skill Identification Results on Sales-Marketing and Finance Areas with Gemma 3}
\label{tab:appendixGemmaArea}
\begin{tabular}{c|lc||c|lc}
\multicolumn{1}{c|}{\textbf{Area}} &
\begin{tabular}[c]{@{}c@{}}\textbf{Prompt}\\\textbf{~Strategy}\end{tabular} & \begin{tabular}[c]{@{}c@{}}\textbf{MUC}\\\textbf{~Partial}\end{tabular} & \multicolumn{1}{c|}{\textbf{Area}} & \begin{tabular}[c]{@{}c@{}}\textbf{Prompt}\\\textbf{~Strategy}\end{tabular} & \begin{tabular}[c]{@{}c@{}}\textbf{MUC}\\\textbf{~Partial}\end{tabular}\\ \hline

\multirow{7}[20]{*}{\rotcell{\mbox{\textbf{Sales-Marketing}}}} 
& Zero-Shot& 0.50 & 
\multirow{7}[10]{*}{\rotcell{\mbox{\textbf{Finance}}}} 
& Zero-Shot& 0.44 \\
& Static Two-Shot  & 0.84 &
    & Static Two-Shot  &    0.84 \\
& Static Six-Shot  & 0.76 &
    & Static Six-Shot   & 0.80 \\
& Static Ten-Shot   & 0.73 &
    & Static Ten-Shot  &  0.76 \\
& Dynamic Two-Shot & 0.69 &          
    & Dynamic Two-Shot & 0.70 \\
& Dynamic Six-Shot  & 0.74 &          
    & Dynamic Six-Shot & 0.76 \\
& Dynamic Ten-Shot &  0.74 &          
    & Dynamic Ten-Shot &  0.76 \\ \hline
\end{tabular}
\end{table}

We conduct a closer analysis of the skill identification results in our two relatively well-sampled sectors, sales-marketing and finance, in order to further discuss the effect of sample size. The training split of our small-sized dataset, although used as the training data for the traditional BERT-based sequence labeler in the skill identification stage, serves instead as an in-context example pool for our LLM-based models, which are introduced as an efficient solution in this low-resource setting, specifically for the dynamic few-shot variants. \Cref{tab:appendixClaudeArea} and \Cref{tab:appendixGemmaArea} provides the results using Claude Sonnet and Gemma models. 

When examining our two relatively well-sampled areas, finance and sales-marketing, we observed that our LLM-based models (Claude and Gemma) achieved scores that are either comparable to or slightly higher than their overall averages under the dynamic ten-shot setup. Specifically, Claude achieved 0.78 in finance and 0.79 in sales-marketing, compared to an overall average of 0.77 (\Cref{tab:skillidentification}). Similarly, Gemma obtained 0.74 in finance and 0.76 in sales-marketing, relative to its overall average of 0.74 (\Cref{tab:skillidentification}). However, since Gemma’s static two-shot variant — which does not employ dynamic sample retrieval and is therefore unaffected by training sample size — also achieved higher results (0.84 for both sectors compared to an overall average of 0.81 (\Cref{tab:skillidentification})), and Claude shows a similar pattern,  these slight improvements in the dynamic ten-shot setting cannot be solely attributed to dataset size.

\subsection{Prompts}

\begin{tcolorbox}[label=appendix:skill_identification_prompt,title=Skill Identification Prompt]
\textbf{Turkish:}
Sana bir ilan vereceğim ve aranan pozisyon için gerekli görülen yetenekleri $<skill\_start>$ ve $<skill\_end>$ etiketleri arasına almanı istiyorum. Açıklama yapma. Metni etiketleri ile birlikte aynı şekilde farklı bir değişiklik yapmadan döndür. Okunan okul ve deneyimler yetenek değildir. 
\\

\textbf{English Translation:}
I will give you a job posting, and I want you to enclose the required skills for the position between the $<skill\_start>$ and $<skill\_end>$ tags. Do not provide any explanations. Return the text exactly as it is with the tags, without making any other changes. Education and experience are not considered skills.
\end{tcolorbox}

\begin{tcolorbox}[title=Skill Linking Prompts - Rerank Prompt]
\label{appendix:skill_linking_prompt-rerank}
\textbf{Turkish:}
Bir insan kaynakları asistanı olarak davran. Sana bir iş ilanında geçen bir yetenek vereceğim. Ayrıca, sana ESCO yetenek veritabanından 10 tane bu yetenekle alakalı olduğunu düşündüğüm yetenekler vereceğim. 
Görevin: Verilen 10 ESCO yeteneğini, verilen yeteneği en kapsayıcı ve en alakalıdan en alakasız olana doğru sıralamak. Bu listeyi oluştururken yeteneklerin yazımını değiştirme, sadece sıralamayı güncelle. Ek açıklamalar yapma.
\\

\textbf{English Translation:}
Act as a human resources assistant. I will give you a skill mentioned in a job posting. Additionally, I will provide you with 10 retrieved skills from the ESCO skill base that I believe are related to this skill.
Your Task: Rank the 10 given ESCO skills from the most comprehensive and relevant to the least relevant in relation to the given skill.
When creating the list:
Do not change the wording of the skills.
Only update the ranking.
Do not provide additional explanations.
\end{tcolorbox}

\begin{tcolorbox}[title=Skill Linking Prompts - Reason + Rerank Prompt]
\label{appendix:skill_linking_prompt-reason_rerank}
\textbf{Turkish:}
Bir insan kaynakları asistanı olarak davran. Sana bir iş ilanından alınmış bir cümle, bu cümlede içinde geçen bir yetenek vereceğim. Ayrıca, sana ESCO yetenek veritabanından 10 tane bu yetenekle alakalı olduğunu düşündüğüm yetenekler vereceğim. Görevin:
 
Sana verilen yeteneği, her bir ESCO yeteneği ile kıyaslamak.
 
Her bir kıyaslama esnasında, her ESCO yeteneğinin, bu ilanda geçen yetenek ile benzerlikleri ve farklılıklarını belirt. Yaptığın açıklamada, sana verilen cümle ve içindeki yetenek ile, kıyaslama yaptığın ESCO yeteneğinin bağlamsal olarak alakalı olup olmadığını açıklayarak anlat.
 
Bütün açıklamaları yaptıktan sonra, verilen 10 ESCO yeteneğini, cümledeki yeteneğe en alakalıdan en alakasıza olacak şekilde sırala.
 
Kıyaslamaları yaparken sana verilen sırayı bozma. Kıyaslama ve açıklamaları yap, daha sonra yeniden sıralanmış bir liste oluştur. Listeyi oluştururken yeteneğin yazımını değiştirme, sadece sıralamayı güncelle.
\\

\textbf{English Translation:}
Act as a human resources assistant. I will provide you with a sentence taken from a job posting and a specific skill mentioned in that sentence. Additionally, I will give you 10 retrieved skills from the ESCO skill base that I believe are related to the given skill. Your task is to compare the given skill with each ESCO skill.

For each comparison, describe the similarities and differences between the ESCO skill and the given skill. In your explanation, clarify whether the ESCO skill is contextually relevant to the given sentence and skill.

After providing all explanations, rank the 10 ESCO skills from most to least relevant to the job posting skill.

While making comparisons, do not change the given order of the skills. First, perform the comparisons and provide explanations. Then, create a reordered list, ensuring that you do not alter the spelling of any skills—only update their ranking.

\end{tcolorbox}
\begin{tcolorbox}[title=Skill Linking Prompts - Causal Reason + Rerank Prompt]
\label{appendix:skill_linking_prompt-causal_reason_rerank}
\textbf{Turkish:}
Bir insan kaynakları asistanı olarak davran. Sana bir iş ilanından alınmış bir yetenek vereceğim. Ayrıca, sana ESCO yetenek veritabanından 10 tane bu yetenekle alakalı olduğunu düşündüğüm yetenekler vereceğim. 

Görevin: Sana verilen yeteneği, her bir ESCO yeteneği ile kıyaslamak.  
Her bir kıyaslama esnasında, kendini iş ilanını veren işveren yerine koy. İşveren olarak, istediğin yetenek ile kıyaslama yaptığın ESCO yeteneğinin uygunluğunu neden ve sonuç olarak kendine açıkla. Yani, bir işveren olarak istediğin yetenekle kıyaslama yaptığın ESCO yeteneğinin NEDEN alakalı olabileceğini düşün, daha sonra istediğin yetenek yerine ESCO yeteneğine sahip olan bir çalışanı işe aldığında, işin gereksinimlerini nasıl yerine getireceği SONUCUNU düşün. Kıyaslamaları yaparken her kıyaslama için neden ve sonuç şeklinde iki açıklamayı da işveren rolü üstlenerek yap. 
 
Bütün açıklamaları yaptıktan sonra, bu 10 ESCO yeteneğini, verilen yeteneği en kapsayıcı ve en alakalıdan en alakasız olana doğru sırala.
Kıyaslamaları yaparken sana verilen sırayı bozma. Kıyaslama ve açıklamaları yap, daha sonra yeniden sıralanmış bir liste oluştur. Listeyi oluştururken yeteneğin yazımını değiştirme, sadece sıralamayı güncelle ve ``Sıralama:'' başlığı altında ver.
\\

\textbf{English Translation:}
Act as a human resources assistant. I will provide you with a skill taken from a job posting. Additionally, I will give you 10 retrieved skills from the ESCO skill base that I believe are related to this skill.

Your task: Compare the given skill with each ESCO skill.
During each comparison, put yourself in the position of the employer who posted the job. As an employer, explain to yourself the relevance of the ESCO skill to the given skill in terms of cause and effect. That is, as an employer, consider WHY the ESCO skill might be related to the given skill, and then think about the OUTCOME of hiring an employee who possesses the ESCO skill instead of the given skill, how they would meet the job requirements. Perform each comparison by providing both the cause and the effect while taking on the role of the employer.

After making all the explanations, rank these 10 ESCO skills from the most comprehensive and relevant to the least relevant in relation to the given skill.
Do not change the given order while making comparisons. Make the comparisons and explanations first, then create a reordered list. When creating the list, do not alter the wording of the skills, just update the ranking and present it under the heading ``Ranking:''.

\end{tcolorbox}

\subsection{Cost Analysis \& Lessons Learned}
\label{sec:appendixcostanalysis}
\begin{table}[ht]
    \centering
    \caption{Average cost per configuration in the Skill Identification stage (Claude Sonnet 3.7) and Reranking stage (GPT-4o), in USD.}
    \label{tab:api-cost}
    \begin{tabular}{ll|ll}
        \multicolumn{2}{c|}{\textbf{Identification}} & \multicolumn{2}{c}{\textbf{Reranking}} \\ \hline
        \textbf{Prompt Type} & \textbf{Cost} & \textbf{Prompt Type} & \textbf{Cost} \\ \hline
        Zero-Shot & \$0.50 & w/o Rerank & - \\
        Two-Shot & \$0.69 & w/ RerankKey Prompt & \$6.39 \\
        Six-Shot & \$1.23 & w/ RerankContext Prompt & \$6.44 \\
        Ten-Shot & \$1.60 & w/ RerankContext-Desc. Prompt & \$7.16 \\
    \end{tabular}
\end{table}

\textbf{Quantitative Comparison.} All experiments were executed either via API access to closed-source LLMs (Claude Sonnet 3.7 for skill identification and GPT-4o for reranking) or on self-hosted GPUs for open-source models (Gemma 3 27B). Based on April 2025 pricing—Claude 3.7 at approximately \$3 per million input and \$15 per million output tokens , GPT-4o at \$5 per million input and \$15 per million output tokens. The average per-configuration cost ranged from \$0.50–\$1.60 for the identification stage with Claude and \$6.39–\$7.16 for reranking with GPT-4o. The latter represents approximately 80-90\% of the total cost.

For comparison, local inference with Gemma 3 27B on 48 GB VRAM GPUs processed a configuration in about 2.5 minutes, yielding an amortized compute cost of \$0.15–\$0.20 per run—approximately 40× cheaper than GPT-4o inference once hardware is available. While API-based execution removes the need for hardware provisioning and ensures reproducibility, the open-source alternative offers a substantially lower marginal cost for repeated experiments. Across all setups, the complete two-stage pipeline remained below \$9 per configuration for proprietary models and below \$0.25 when self-hosted.

\textbf{Cost–Performance Trade-offs.} Our experiments (\Cref{tab:skillextractionresultswithstandarderrors}) show that the majority of performance improvements occur between zero-shot and two-shot configurations, whereas additional examples yield diminishing returns. In the embedding similarity setup, skill identification with static two-shot and dynamic six-shot prompts achieved average end-to-end scores of 0.55 and 0.54 under the w/ RerankContext-Description configuration, closely approaching the dynamic ten-shot configuration (0.56) while incurring roughly half the cost. Likewise, the RerankKey and RerankContext prompts performed within 0.01 end-to-end score of RerankContext-Desc yet required around 10\% fewer costs. These findings suggest that researchers can preserve experimental soundness by (i)~adopting mid-range few-shot settings, (ii) employing medium-complexity reranking prompts, and (iii) reserving full two-stage runs for final evaluation. Following these guidelines would reduce the total expenditure by approximately 20\% with some loss in score.

\textbf{Lessons Learned for Cost-Effective Experimentation.} Several practices can be adopted to reduce computational expenses in future experiments.
\begin{enumerate}
    \item Stage separation: conducting exploratory analyses using identification-only runs before invoking reranking allowed faster iteration and inexpensive debugging.
    \item Caching and reuse: intermediate skill span - ESCO retrieved skill lists could be reused across reranker variants, avoiding repeated inference on identical text.
    \item Dynamic prompting: our experiments demonstrate that dynamic prompting achieves results that are consistently the same as or better than static prompting across equivalent few-shot settings, without increasing token count. This finding suggests that testing static variants of similar systems offers limited additional value, as dynamic prompting provides comparable effectiveness with no cost penalty.
    \item Lightweight encoders: adopting more lightweight approaches such as ConTeXT-match \citep{Decorte_2025} can deliver near-LLM extraction quality at a fraction of the computational cost, offering an efficient alternative for large-scale deployment.
    \item Parameter-efficient fine-tuning: While this study relies exclusively on prompting and does not involve any fine-tuning, this choice is partly due to the low-resource nature of our setting, both in terms of language (Turkish) and available training data. Fine-tuning represents a viable alternative to closed-source LLMs when sufficient data and resources are available.In particular, adapting open-source models through PEFT methods such as LoRA \citep{hu2021loralowrankadaptationlarge}, QLoRA \citep{dettmers2023qloraefficientfinetuningquantized}, or adapter-based tuning could improve performance while enabling cost-effective, fully local deployment. By updating only a small fraction of model parameters, this approach could significantly reduce long-term inference costs without sacrificing quality, making it a practical option for future work in richer-resource scenarios.
\end{enumerate}

Together, these approaches represent feasible strategies for scaling skill-extraction research while maintaining experimental rigor and reproducibility in resource-constrained environments.

\subsection{Additional Results}
\label{sec:additionalresults}
\begin{table}[ht]
\centering
\caption{Skill Identification Results for Gemma 3 and Claude Sonnet 3.7: averaged over Three Runs with Standard Errors
(Mean $\pm$ SE)}
\label{tab:meanskillidentification}
\begin{tabular}{c|lcc}
\multicolumn{1}{c|}{\textbf{LLM}} &
\begin{tabular}[c]{@{}c@{}}\textbf{Prompt}\\\textbf{~Strategy}\end{tabular} &
\begin{tabular}[c]{@{}c@{}}\textbf{CoNLL}\\\textbf{~F1}\end{tabular} & \begin{tabular}[c]{@{}c@{}}\textbf{MUC}\\\textbf{~Partial}\end{tabular} \\ \hline
 
\multirow{7}[10]{*}{\rotcell{\mbox{\begin{tabular}[c]{@{}c@{}}\textbf{Claude}\\\textbf{~Sonnet 3.7}\end{tabular}}}} 
& Zero-Shot& $0.25 \pm _{0.003}$ & $0.63 \pm _{0.005}$\\
& Static Two-Shot & $0.23 \pm_{0.001}$	& $0.75 \pm_{0.001}$\\
& Static Six-Shot & $0.16 \pm_{0.001}$ & $0.75 \pm_{0.001}$ \\
& Static Ten-Shot & $0.26 \pm _{0.001}$ & $0.76 \pm _{0.001}$ \\
& Dynamic Two-Shot &  $0.48 \pm _{0.003}$	& $0.73 \pm _{0.001}$ \\    
& Dynamic Six-Shot & $0.53 \pm_{0.002}$ & $0.75 \pm _{0.001}$ \\
& Dynamic Ten-Shot & $0.57 \pm_{0.001}$ & $0.78 \pm_{0.001}$ \\   \hline
\multirow{7}[10]{*}{\rotcell{\mbox{\textbf{Gemma 3}}}} 
& Zero-Shot& $0.19 \pm_{0.000}$ & $0.44 \pm_{0.001}$\\
& Static Two-Shot & $0.17 \pm_{0.002}$	& $0.81 \pm_{0.004}$ \\
& Static Six-Shot & $0.25 \pm_{0.001}$ & $0.77 \pm_{0.001}$ \\
& Static Ten-Shot & $0.32 \pm _{0.002}$ & $0.72 \pm _{0.003}$ \\
& Dynamic Two-Shot & $0.42 \pm _{0.003}$ & $0.69 \pm _{0.001}$ \\
& Dynamic Six-Shot & $0.52 \pm_{0.001}$ & $0.73 \pm_{0.001}$ \\  
& Dynamic Ten-Shot & $0.52 \pm_{0.001}$ & $0.75 \pm_{0.001}$ \\ \hline
\end{tabular}
\end{table} 

\begin{table}[ht]
\centering
\caption{Results for End-to-End Skill Extraction: Identification with Gemma 3, Reranking by GPT-4o w/ Causal Reasoning with Standard Errors}
\label{tab:skillextractionresultswithstandarderrorsgemma}
\begin{tabular}{c|l|cccc}

\multirow{2}[0]{*}{\mbox{
\begin{tabular}[c]{@{}c@{}}\textbf{\textbf{Retrieval }}\\\textbf{\textbf{Methods}}\end{tabular}
} }

 & \multirow{2}[0]{*}{\mbox{
\begin{tabular}[c]{@{}c@{}}\textbf{\textbf{Identification }}\\\textbf{\textbf{Methods}}\end{tabular}
} } &  \multicolumn{4}{c}{\textbf{Reranking Methods}} \\ 
 &  &  \begin{tabular}[c]{@{}c@{}}\textbf{\textbf{w/o }}\\\textbf{\textbf{~Rerank}}\end{tabular}
 & \begin{tabular}[c]{@{}c@{}}\textbf{\textbf{w/ RerankKey}}\\\textbf{\textbf{~Prompt}}\end{tabular} & \begin{tabular}[c]{@{}c@{}}\textbf{\textbf{w/ RerankContext }}\\\textbf{\textbf{Prompt}}\end{tabular} &
\begin{tabular}[c]{@{}c@{}}\textbf{\textbf{w/ RerankContext-}}\\\textbf{\textbf{~Desc. Prompt}}\end{tabular}
\\ \hhline{------}
\multirow{8}[15]{*}{\rotcell {\mbox{
\begin{tabular}[c]{@{}c@{}}\textbf{\textbf{Fuzzy }}\\\textbf{\textbf{Matching}}\end{tabular}}}} 
& SSL via Bert~ & $0.17\pm_{0.000}$ & $0.29\pm_{0.000}$ & $0.30\pm_{0.000}$ & $0.29\pm_{0.000}$   \\
 & Zero-Shot & $0.15\pm_{0.000}$ & $0.26\pm_{0.003}$ & $0.26\pm_{0.003}$ & $0.26\pm_{0.000}$   \\
 & Static Two-Shot & $0.14\pm_{0.000}$ & $0.25\pm_{0.000}$ & $0.26\pm_{0.003}$ & $0.24\pm_{0.003}$   \\
 & Static Six-Shot & $0.14\pm_{0.000}$ & $0.22\pm_{0.003}$ & $0.23\pm_{0.005}$ & $0.23\pm_{0.005}$   \\
 & Static Ten-Shot & $0.14\pm_{0.000}$ & $0.27\pm_{0.000}$ & $0.27\pm_{0.000}$ & $0.27\pm_{0.003}$   \\
 & Dynamic Two-Shot & $0.18\pm_{0.003}$ & $0.30\pm_{0.005}$ & $0.30\pm_{0.000}$ & $0.30\pm_{0.003}$   \\
 & Dynamic Six-Shot & $0.20\pm_{0.000}$ & $0.32\pm_{0.000}$ & $0.32\pm_{0.000}$ & $0.32\pm_{0.003}$   \\
 & Dynamic Ten-Shot& $0.20\pm_{0.003}$ & $0.33\pm_{0.003}$ & $0.33\pm_{0.000}$ & $0.32\pm_{0.003}$   \\ \hhline{------}
\multirow{8}[18]{*}{\rotcell {\mbox{
\begin{tabular}[c]{@{}c@{}}\textbf{\textbf{Embedding }}\\\textbf{\textbf{Similarity}}\end{tabular}}}} 
& SSL via Bert~ & $0.19\pm_{0.000}$ & $0.35\pm_{0.000}$ & $0.36\pm_{0.000}$ & $0.37\pm_{0.000}$   \\
 & Zero-Shot & $0.33\pm_{0.005}$ & $0.43\pm_{0.005}$ & $0.45\pm_{0.003}$ & $0.45\pm_{0.005}$   \\
 & Static Two-Shot & $0.38\pm_{0.000}$ & $0.52\pm_{0.007}$ & $0.52\pm_{0.003}$ & $0.54\pm_{0.000}$   \\
 & Static Six-Shot & $0.37\pm_{0.005}$ & $0.52\pm_{0.003}$ & $0.53\pm_{0.007}$ & $0.53\pm_{0.005}$   \\
 & Static Ten-Shot & $0.38\pm_{0.000}$ & $0.52\pm_{0.003}$ & $0.53\pm_{0.005}$ & $0.53\pm_{0.003}$   \\
 & Dynamic Two-Shot & $0.38\pm_{0.003}$ & $0.52\pm_{0.003}$ & $0.52\pm_{0.005}$ & $0.53\pm_{0.007}$   \\
 &  Dynamic Six-Shot & $0.41\pm_{0.000}$ & $0.54\pm_{0.005}$ & $0.54\pm_{0.005}$ & $0.53\pm_{0.003}$   \\
 & Dynamic Ten-Shot & $0.41\pm_{0.000}$ & $0.55\pm_{0.005}$ & $0.54\pm_{0.007}$ & $0.55\pm_{0.007}$   \\ \hhline{------}
\end{tabular}

\end{table}

\begin{table}[ht]
\centering
\caption{Results for End-to-End Skill Extraction:  Identification with GPT-4o,  Reranking by GPT-4o w/ Causal Reasoning}
\label{tab:skillextractionresultswithgpt4o}
\begin{tabular}{c|l|cccc}
\multirow{2}[0]{*}{\mbox{
\begin{tabular}[c]{@{}c@{}}\textbf{\textbf{Retrieval }}\\\textbf{\textbf{Methods}}\end{tabular}}}
 & \multirow{2}[0]{*}{\mbox{
\begin{tabular}[c]{@{}c@{}}\textbf{\textbf{Identification }}\\\textbf{\textbf{Methods}}\end{tabular}
} } &  \multicolumn{4}{c}{\textbf{Reranking Methods}}\\

 &  &
 \begin{tabular}[c]{@{}c@{}}\textbf{\textbf{w/o }}\\\textbf{\textbf{~Rerank}}\end{tabular}
 & \begin{tabular}[c]{@{}c@{}}\textbf{\textbf{w/ RerankKey}}\\\textbf{\textbf{~Prompt}}\end{tabular} & \begin{tabular}[c]{@{}c@{}}\textbf{\textbf{w/ RerankContext }}\\\textbf{\textbf{Prompt}}\end{tabular} &
\begin{tabular}[c]{@{}c@{}}\textbf{\textbf{w/ RerankContext-}}\\\textbf{\textbf{~Desc. Prompt}}\end{tabular}
\\ \hhline{------}
\multirow{8}[15]{*}{\rotcell {\mbox{
\begin{tabular}[c]{@{}c@{}}\textbf{\textbf{Fuzzy }}\\\textbf{\textbf{Matching}}\end{tabular}}}} 
& SSL via Bert~ &  0.17 & 0.29  & 0.30 & 0.29 \\
 & Zero-Shot & 0.11 & 0.18 & 0.19 & 0.19 \\
 & Static Two-Shot & 0.12 & 0.19 & 0.19 & 0.19 \\
 & Static Six-Shot & 0.12 & 0.19 & 0.29 & 0.19 \\
 & Static Ten-Shot & 0.13 & 0.22 & 0.22 & 0.22 \\
 & Dynamic Two-Shot & 0.16 & 0.26 & 0.27 & 0.26\\
 & Dynamic Six-Shot & 0.16 & 0.28 & 0.27 & 0.28\\
 & Dynamic Ten-Shot & 0.18 & 0.30 & 0.30 & 0.30 \\ \hhline{------}
\multirow{8}[18]{*}{\rotcell {\mbox{
\begin{tabular}[c]{@{}c@{}}\textbf{\textbf{Embedding }}\\\textbf{\textbf{Similarity}}\end{tabular}}}} 
& SSL via Bert~ & 0.19 & 0.35 & 0.36 & 0.37 \\
 & Zero-Shot & 0.32 & 0.43 & 0.43 & 0.43 \\
 & Static Two-Shot & 0.32 & 0.41 & 0.40 & 0.42 \\
 & Static Six-Shot & 0.32 & 0.44 & 0.42 & 0.44 \\
 & Static Ten-Shot & 0.36 & 0.48 & 0.47 & 0.48 \\
 &Dynamic Two-Shot & 0.35 & 0.49 & 0.49 & 0.52\\
 &  Dynamic Six-Shot & 0.34 & 0.50 & 0.50 & 0.50\\
 & Dynamic Ten-Shot & 0.38 & \textbf{0.55} & 0.53 & 0.52 \\ \hhline{------}
\end{tabular}
\label{Performance of end-to-end Skill Extraction}
\end{table}

\begin{table}[ht]
\caption{Color-Coded Human Evaluations}
\label{appendix:humeval}
    \centering
   \begin{tabular}{lllllp{8cm}}
        & \textbf{\# of Extracted} & \multicolumn{4}{l}{\textbf{Human Evaluation}}                                                                                                                                                     \\
\textbf{JobID}   & \textbf{Skills}          &\textbf{Red}     & \textbf{Blue}    & \textbf{Gray}   & \textbf{Main Comment}                \\\hline
120 & 17              & 9       & 6       & 2      & ``Power BI'' listed in Job Ads but was not extracted, ``fiyat ürün'' (price product) needs more context.  \\
184 & 3               & 3       & 0       & 0      & n/a \\
213 & 17              & 11      & 6       & 0      & ``Logo Muhasebe'' is a critical skill which was not extracted. This, can be labeled as critical omission of system. \\
109 & 19              & 12      & 5       & 2      & ``Level of English'' Language can be extracted too, for example ``Advanced English'' is stated in job ads, ``Plan'' is an ambiguous extraction as  it needs more extra context or keyword extraction \\
273 & 15              & 11      & 3       & 1      & This job does not require business analysis as skill, mismatch of skill and position. \\
289 & 8               & 5       & 1       & 2      & Since the position did not require leadership competency, supervisory skills seemed unnecessary. \\
250 & 11              & 7       & 1       & 3      & They gray skills are not common or expected from candidate profile \\
220 & 17              & 10      & 2       & 5      & As example ``work in team'' does not reflect a skill, but it is a ``competency''. It may not necessarily be found in the candidate's resume. \\
320 & 2               & 1       & 1       & 0      & n/a \\
135 & 11              & 8       & 2       & 1      & n/a \\
275 & 9               & 9       & 0       & 0      & n/a \\
176 & 4               & 4       & 0       & 0      & n/a \\
94 & 6               & 5       & 1       & 0      & n/a \\
234 & 9               & 8       & 1       & 0      & n/a \\
57 & 10              & 6       & 0       & 4      & Example of gray skills: ``sell products'' labeled as skills but it is mainly considered as task or responsibility in Turkish labor market. \\
138 & 4               & 1       & 3       & 0      & n/a \\
52 & 13              & 11      & 1       & 1      & n/a \\
200 & 4               & 3       & 1       & 0      & n/a \\
297 & 7               & 7       & 0       & 0      & n/a \\
153 & 7               & 5       & 1       & 1      & n/a \\
295 & 8               & 5       & 3       & 0      & Some skills are not critical or optional in Turkish labor market, example: ``purchase supplies'' in this extracted part. \\
143 & 2               & 2       & 0       & 0      & n/a \\
264 & 27              & 26      & 1       & 0      & n/a \\
Total   & 230             & 169     & 39      & 22     &   \\
Perc.   &                 & 73\% & 17\% & 10\% &                                                                                                                                                            
\end{tabular}
\end{table}

\end{document}